\documentclass[runningheads]{llncs}
\usepackage[T1]{fontenc}
\usepackage{graphicx}
\usepackage{cite}
\usepackage{amsmath,amssymb,amsfonts}
\usepackage{algorithmic}
\usepackage{graphicx}
\usepackage{textcomp}
\usepackage{xcolor}
\usepackage{tikz}
\usepackage{booktabs}
\usepackage{multirow}
\usetikzlibrary{arrows.meta, positioning}
\usepackage{float}
\usepackage{placeins}
\usepackage{tikz}
 \usepackage{subcaption} 
 \usepackage{adjustbox}
\usetikzlibrary{shapes.geometric, arrows.meta, positioning}
\usepackage[table]{xcolor}
\usepackage{longtable}
\usepackage{graphicx}
\usepackage{booktabs}
\usepackage{pgfplots}
\usepgfplotslibrary{polar}
\pgfplotsset{compat=1.18} 
\usepackage{longtable}
\usepackage{multirow}
\usepackage{array} 
\usepackage{threeparttable} 
\usepackage{microtype}

\usepackage{xcolor}
\usepackage[table]{xcolor}
\definecolor{darkgreen}{RGB}{0,153,0}
\definecolor{darkred}{RGB}{192,0,0}
\definecolor{green1}{RGB}{222,255,226}
\definecolor{orange1}{RGB}{255,207,156}
\definecolor{red1}{RGB}{240,180,180}

\usepackage[colorlinks,urlcolor=blue,linkcolor=blue,citecolor=blue]{hyperref}

\newcolumntype{C}[1]{>{\centering\arraybackslash}m{#1}}
\usepackage{tabularx} 

\usepackage{placeins}
\begin{document}

%
\title{From Visual Feedback to Textual Reviews: A Multi-Agent Vision-Language Framework for Image-Grounded Review Assistance}
\titlerunning{From Visual Feedback to Textual Reviews ...}
%
\author{Ayush Bansal\inst{1}\orcidID{0000-1111-2222-3333} \and
Utsav Kumar Nareti\inst{1}\orcidID{2222--3333-4444-5555} \and
Kumari Priya\inst{1}\orcidID{1111-2222-3333-4444} \and
Saba Akram \inst{1}\orcidID{2222--3333-4444-5555} \and
Chandranath Adak\inst{1}\orcidID{2222--3333-4444-5555} \and
Muhammad Saqib\inst{3}\orcidID{2222--3333-4444-5555}}
\authorrunning{Ayush et al.}
%
\institute{Department of Computer Science and Engineering, Indian Institute of Technology, Patna, Bihar 801106, India \and
National Collection and Marine Infrastructure, CSIRO, Canberra, ACT 2601, Australia
}

\author{Utsav Kumar Nareti\inst{1}\orcidID{0000-0002-1578-371X} \and
Ayush Bansal\inst{1}\orcidID{0009-0006-3232-2647} \and 
Kumari Priya\inst{1}\orcidID{0009-0001-7078-0490} \and 
Chandranath Adak\inst{1}\orcidID{0000-0002-9085-2770} \and
Soumi Chattopadhyay \inst{2}\orcidID{0000-0002-9231-4087} \and
Muhammad Saqib \inst{3}\orcidID{0000-0003-4374-0888} \and
Saeed Anwar \inst{4}\orcidID{0000-0002-0692-8411}
} 
\authorrunning{U. K. Nareti et al.}
\institute{
Dept. of CSE, Indian Institute of Technology Patna, India-801106 \and 
Dept. of CSE, Indian Institute of Technology Indore, India-453552 \and
NCMI, CSIRO, Australia-2601 \and 
University of Western Australia, Australia-6009\\
\email{ \{utsav\_2221cs28,chandranath\}@iitp.ac.in}
}

\maketitle              

\begin{abstract}
Visual feedback in the form of user-uploaded images and videos is becoming increasingly common in e-commerce platforms because it provides authentic evidence of product quality, defects, packaging conditions, and real-world usage. However, visual feedback alone often lacks the contextual explanations and subjective opinions necessary for informed decision-making, while many users provide limited textual feedback due to the effort required to compose detailed reviews. To bridge this gap, we introduce \emph{image-grounded review assistance}, a novel task that aims to generate editable review drafts from user-uploaded product images. Unlike conventional image captioning, which focuses on objective visual description, the proposed task requires product-specific understanding, sentiment estimation, and evidence-driven review composition under challenging real-world conditions, including degraded image quality, excessive zoom-in, target ambiguity, and partial product visibility. We propose a multi-agent vision-language framework consisting of four specialised roles: product grounding, visual sentiment estimation, visual evidence generation, and review synthesis. The framework employs explicit intermediate representations, including product entities, predicted ratings, and evidence summaries, to improve interpretability and visual grounding. Experiments on a curated subset of the Amazon Reviews Electronics dataset demonstrate the feasibility of generating coherent, product-aware, and sentiment-aware review drafts from visual feedback. To the best of our knowledge, this is the first study to formulate image-grounded review assistance as a multi-agent vision-language reasoning problem, providing a practical step toward AI-assisted review authoring in e-commerce systems.

\keywords{
Image-Grounded Review Assistance \and 
Multi-Agent Vision-Language Systems \and 
E-Commerce Analytics \and 
Review Generation}

\end{abstract}

\section{Introduction}
\label{sec:intro}

The rapid growth of online marketplaces has transformed consumer behaviour by making product discovery, comparison, and purchasing more accessible, a trend further accelerated by the global pandemic \cite{intro_covid_impact}. As competition among platforms intensifies, customer trust, transparency, and satisfaction have become critical factors for sustainable growth, with customer reviews playing a central role in informing purchase decisions, improving recommendations, and assessing seller performance \cite{intro_trust,intro_review}. Consequently, modern e-commerce platforms increasingly encourage multimodal feedback, including ratings, text, images, and videos \cite{intro_image,intro_video}. In recent years, users have shown a growing preference for visual feedback because uploading images or videos of damaged products, defective components, incorrect packaging, or real-world usage scenarios is often faster and more convenient than writing detailed reviews. Such visual feedback provides authentic evidence of product appearance, quality, defects, and usage conditions, often capturing information absent from brief textual comments \cite{intro-vinyal}. Prior studies have further shown that visual information significantly influences consumer perception and can support tasks such as product category recognition, review helpfulness analysis, and review score prediction \cite{intro_image,intro_video,adakdeep}.

Despite its practical value, visual feedback alone is often insufficient for conveying a complete user experience. A product image may reveal what happened, but rarely explains why a user is satisfied or dissatisfied. Moreover, user-generated images collected in realistic e-commerce environments are highly unconstrained and frequently suffer from degraded quality, motion blur, poor illumination, excessive zoom-in, target ambiguity, partial product visibility, cluttered backgrounds, and damaged-product-centric views. These factors make visual evidence difficult to interpret and limit its usefulness for both consumers and platform operators. Consequently, although users increasingly upload visual feedback, the absence of accompanying textual explanations often leaves important contextual information unspecified. 

Traditional review mining research has primarily focused on textual reviews, including sentiment classification \cite{pang2002thumbs}, aspect-based opinion mining \cite{hu2004mining}, and review helpfulness prediction \cite{kim2006automatically}. While effective, these approaches assume the availability of sufficiently detailed textual feedback. In practice, however, many users provide only short comments or omit textual reviews altogether. Recent studies have demonstrated that review images provide complementary information and significantly influence consumer perception and review helpfulness \cite{kubler2024effect}. Visual sentiment analysis further suggests that images may contain affective cues that can support opinion inference \cite{you2015robust}. Nevertheless, transforming visual feedback into meaningful user-oriented textual feedback remains largely unexplored. 

A natural solution might be to employ image captioning models. However, image-grounded review assistance is fundamentally different from conventional image captioning and visual description generation \cite{show,li2022blip,llava}. Image captioning aims to objectively describe visible content, whereas review assistance requires generating subjective, opinion-oriented feedback that reflects potential user experiences. Furthermore, captions primarily describe \emph{what} appears in an image, while review assistance must reason about \emph{how} the observed visual evidence relates to product quality, satisfaction, or dissatisfaction. This requires product-specific understanding, sentiment estimation, and evidence aggregation beyond conventional object recognition and scene description. Therefore, effective review assistance demands an interpretable reasoning process that explicitly connects product identity, visual evidence, and inferred sentiment.

Motivated by these observations, we introduce \emph{image-grounded review assistance}, a task that aims to automatically generate an editable review draft from user-uploaded product images. The goal is not to replace genuine customer opinions, but to assist users in documenting their experiences more efficiently by providing an initial textual draft that can be freely modified, refined, or rewritten before submission. Unlike prior studies on image captioning, visual sentiment analysis, review helpfulness prediction, and multimodal review understanding, our objective is neither to describe image content nor to analyse existing reviews. Instead, we seek to bridge the gap between increasingly prevalent visual feedback and traditional textual feedback in e-commerce systems. 
To address this problem, we propose a multi-agent vision-language framework consisting of four specialised stages: product grounding, visual sentiment estimation, visual evidence generation, and review synthesis. The modular architecture provides interpretable intermediate representations and enables explicit reasoning over product identity, sentiment, and visual evidence before generating the final review draft. We evaluate the proposed framework on the Amazon Reviews Electronics dataset \cite{amazonDataset}. The experimental results demonstrate the feasibility of generating coherent, visually grounded, and sentiment-aware review drafts from product images while highlighting the challenges associated with inferring subjective user experiences from visual evidence alone. 
To the best of our knowledge, this is the first work to formulate image-grounded review assistance as a multi-agent vision-language reasoning problem that jointly performs product grounding, sentiment estimation, visual evidence extraction, and review synthesis within a unified and interpretable framework. 
Our main \textbf{contributions} are summarized as follows:

\textit{(i)} We introduce \emph{image-grounded review assistance}, a new task that aims to generate editable textual review drafts from user-uploaded product images, bridging the gap between increasingly prevalent visual feedback and traditional textual reviews in e-commerce platforms. To the best of our knowledge, this is the first study to formulate this problem within a vision-language reasoning framework.


\textit{(ii)}  We propose a multi-agent vision-language framework that decomposes the task into four interpretable stages: product grounding, visual sentiment estimation, visual evidence generation, and review synthesis. The framework incorporates explicit intermediate representations, including product entities, predicted ratings, and visual evidence summaries, to improve interpretability, controllability, and visual grounding.

\textit{(iii)} We conduct extensive experiments on the Amazon Reviews Electronics dataset and demonstrate the feasibility of generating coherent, product-aware, and sentiment-aware review drafts from visual feedback, while providing detailed analyses of the strengths and limitations of the proposed approach.

The remainder of this paper is organized as follows. Section \ref{sec:dataset} describes the dataset, 
Section \ref{sec:method1} presents the proposed methodology, 
Section \ref{4sec:exp} reports the experimental results, and 
Section \ref{5sec:discussion_conclusion} concludes the paper.








\section{Dataset and Challenges}
\label{sec:dataset}
\definecolor{coffee}{HTML}{4E6FDA}
\definecolor{garbage}{HTML}{ED4C3B}
\definecolor{freeze}{HTML}{F0CD38}
\definecolor{washer}{HTML}{45C35D}
\definecolor{oven}{HTML}{EC7D0E}
\definecolor{range}{HTML}{4CC7F2}
\definecolor{humid}{HTML}{8F85CD}
\definecolor{commercial}{HTML}{CA88A6}

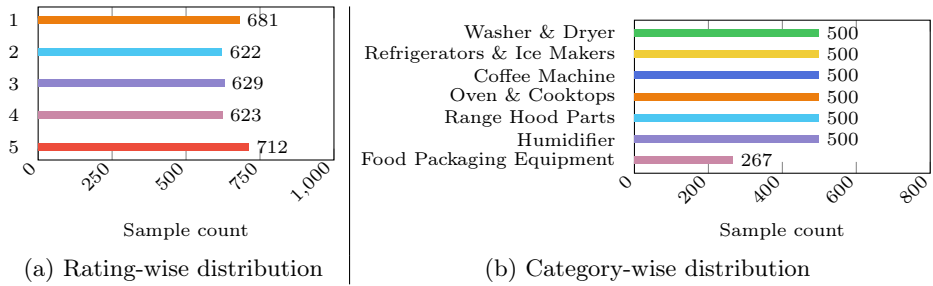
\begin{figure}[!b]
    \centering

\begin{tabular}{c|c}
\scriptsize
        \begin{tikzpicture}
        \begin{axis}[
            width=0.45\textwidth,
            xbar,
            bar width=3pt,
            xmin=0,
            xmax=1000,
            xtick={0,250,500,750,1000},
            ytick={
                5,4,3,2,1
            },
            symbolic y coords={
                5,4,3,2,1
            },
            y=12pt, 
            yticklabel style={align=right},
            nodes near coords align={horizontal},
            nodes near coords,
            nodes near coords style={text=black, anchor=west},
            xticklabel style={rotate=45, anchor=east},   
            xlabel={Sample count},
            enlarge y limits,
            ytick style={draw=none}
        ]
        \addplot+[xbar, fill=garbage, draw opacity=0, bar shift=0pt] coordinates {(712,5)};
        \addplot+[xbar, fill=commercial, draw opacity=0, bar shift=0pt] coordinates {(623,4)};
        \addplot+[xbar, fill=humid, draw opacity=0, bar shift=0pt] coordinates {(629,3)};
        \addplot+[xbar, fill=range, draw opacity=0, bar shift=0pt] coordinates {(622,2)};
        \addplot+[xbar, fill=oven, draw opacity=0, bar shift=0pt] coordinates {(681,1)};
        \end{axis}
        \end{tikzpicture}
&
        \begin{tikzpicture}
        \scriptsize
        \begin{axis}[
            width=0.45\textwidth,
            xbar,
            bar width=3pt,
            xmin=0,
            xmax=800,
            xtick={0,200,400,600,800},
            ytick={
                Food Packaging Equipment,
                Humidifier,
                Range Hood Parts,
                Oven \& Cooktops,
                Coffee Machine,
                {Refrigerators \& Ice Makers},
                Washer \& Dryer
            },
            symbolic y coords={
                Food Packaging Equipment,
                Humidifier,
                Range Hood Parts,
                Oven \& Cooktops,
                Coffee Machine,
                {Refrigerators \& Ice Makers},
                Washer \& Dryer
            },
            y=8pt, 
            yticklabel style={align=right,},
            nodes near coords align={horizontal},
            nodes near coords,
            nodes near coords style={text=black, anchor=west},
            xticklabel style={rotate=45, anchor=east},   
            xlabel={Sample count},
            enlarge y limits,
            ytick style={draw=none}
        ]
        \addplot+[xbar, fill=commercial, draw opacity=0, bar shift=0pt] coordinates {(267,Food Packaging Equipment)};
        \addplot+[xbar, fill=humid, draw opacity=0, bar shift=0pt] coordinates {(500,Humidifier)};
        \addplot+[xbar, fill=range, draw opacity=0, bar shift=0pt] coordinates {(500,Range Hood Parts)};
        \addplot+[xbar, fill=oven, draw opacity=0, bar shift=0pt] coordinates {(500,Oven \& Cooktops)};
        \addplot+[xbar, fill=coffee, draw opacity=0, bar shift=0pt] coordinates {(500,Coffee Machine)};
        \addplot+[xbar, fill=freeze, draw opacity=0, bar shift=0pt] coordinates {(500,{Refrigerators \& Ice Makers})};
        \addplot+[xbar, fill=washer, draw opacity=0, bar shift=0pt] coordinates {(500,Washer \& Dryer)};
        \end{axis}
        \end{tikzpicture}
    \\
    (a) Rating-wise distribution & (b) Category-wise distribution
\end{tabular}
\caption{Dataset distribution}
\label{fig:hr_dataset_distribution}
\end{figure}

We engage the Amazon Reviews Electronics dataset \cite{amazonDataset}, a comprehensive repository containing user-uploaded product images, textual reviews, and corresponding sentiment ratings. Since our primary objective is to generate opinion-oriented reviews grounded exclusively in visual perception, we require high-quality and relevant image-text pairs. Consequently, we employ a rigorous curation protocol to construct a high-fidelity benchmark subset for image-grounded reasoning. First, we enforce strict visual-textual alignment constraints by systematically filtering out samples that lack valid, user-uploaded product images. We prioritize examples where the visual content exhibits a strong correlative link with the user's feedback, ensuring that the underlying task remains focused on genuine visual grounding. Furthermore, to facilitate a robust and systematic evaluation, we reorganize the dataset's native, highly fragmented product categories into a more cohesive semantic taxonomy. The original fine-grained categories are aggregated into seven categories based on semantic similarity, shared visual attributes, and sample density. This process yields a curated subset comprising 3267 examples. The detailed distribution of this finalized dataset is illustrated in Fig. \ref{fig:hr_dataset_distribution}.

\begin{figure}[!t]
\centering
\newcommand{\myImgW}{0.1\linewidth}
\newcommand{\myImgH}{1cm}
\setlength{\tabcolsep}{1pt} 
\renewcommand{\arraystretch}{1.3} 

\begin{adjustbox}{width=0.6\linewidth} 
\footnotesize 
\begin{tabular}{c|c|c|c|c}

\hline 
&  & & & \\ [\dimexpr-\normalbaselineskip+0.5pt]
\includegraphics[width=\myImgW, height=\myImgH]{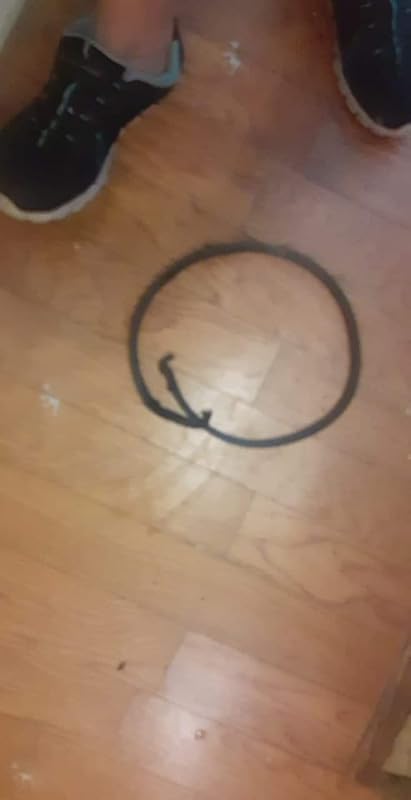} &
\includegraphics[width=\myImgW, height=\myImgH]{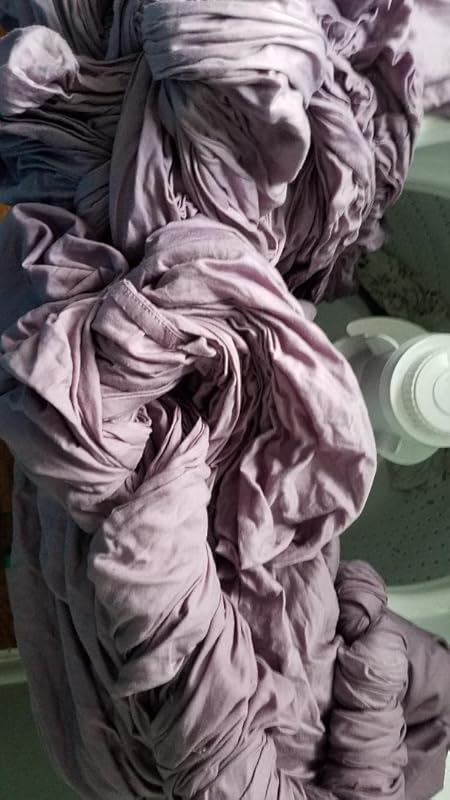} &
\includegraphics[width=\myImgW, height=\myImgH]{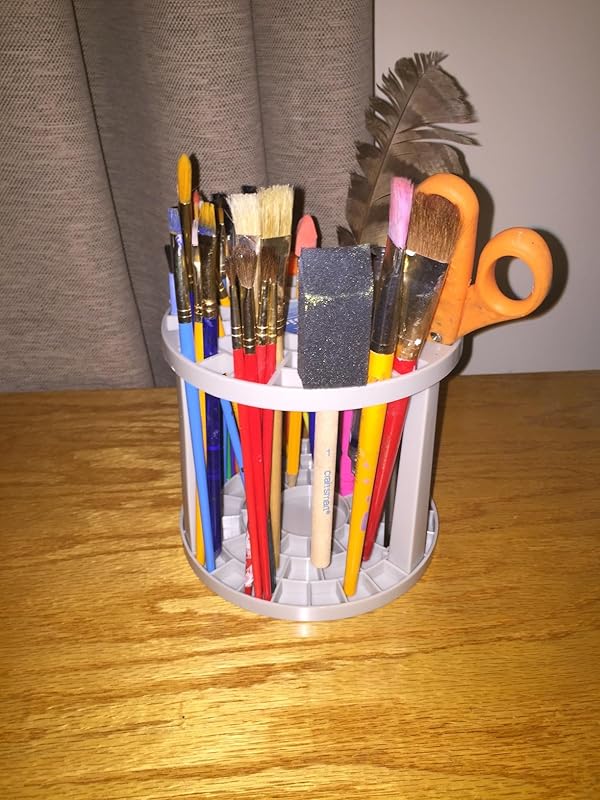} &
\includegraphics[width=\myImgW, height=\myImgH]{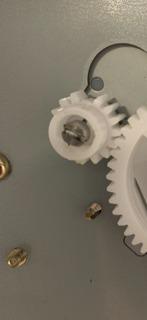} &
\includegraphics[width=\myImgW, height=\myImgH]{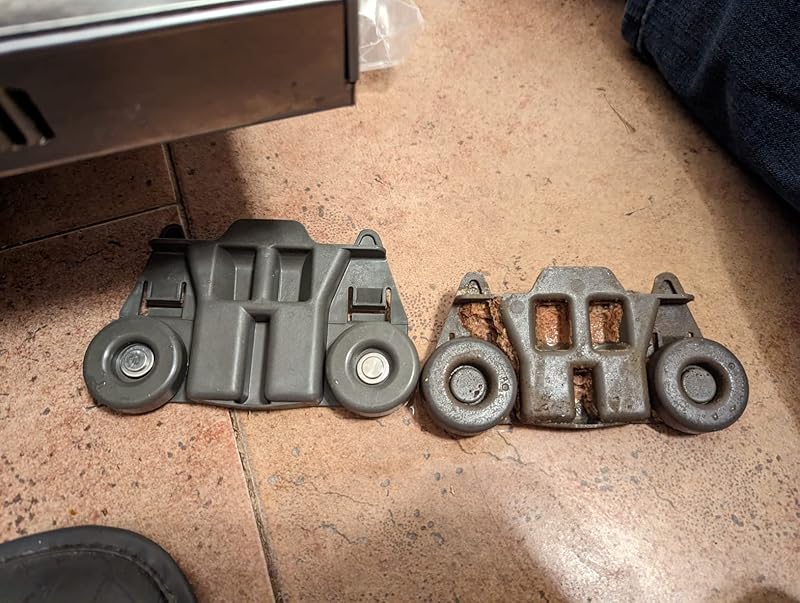} \\

\hline 
&  & & & \\ [\dimexpr-\normalbaselineskip+0.5pt]

\includegraphics[width=\myImgW, height=\myImgH]{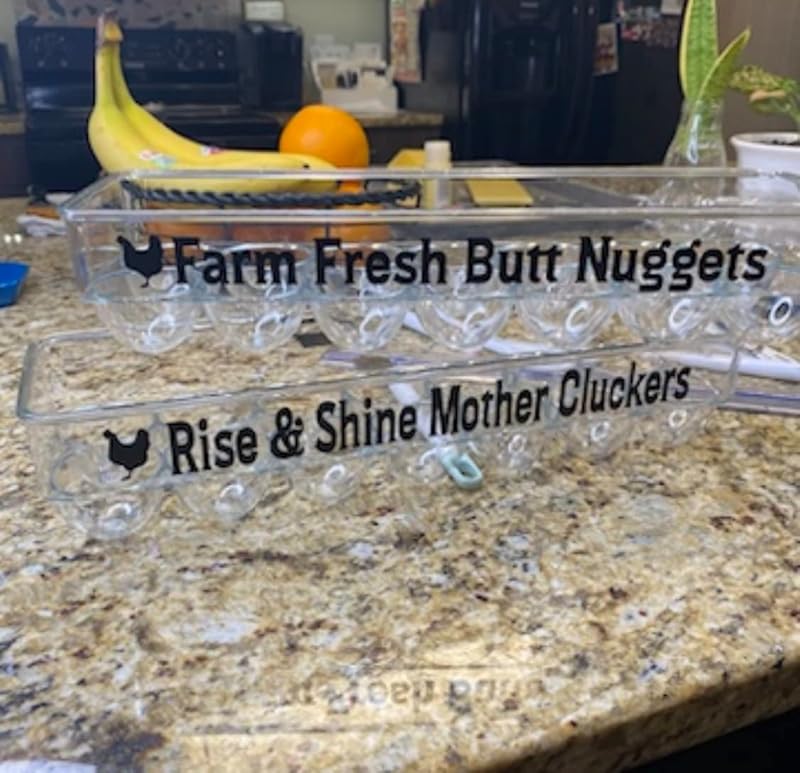} &
\includegraphics[width=\myImgW, height=\myImgH]{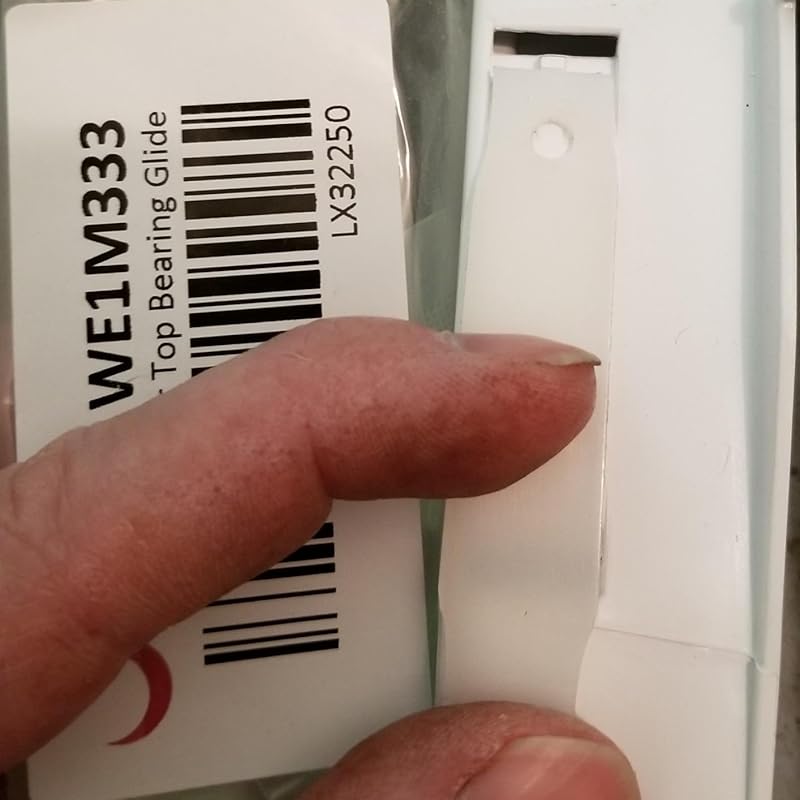} &
\includegraphics[width=\myImgW, height=\myImgH]{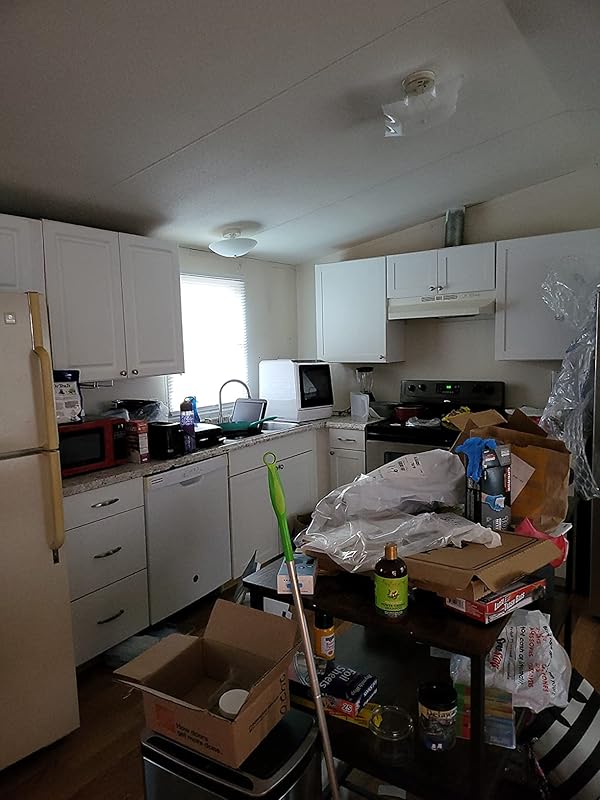} &
\includegraphics[width=\myImgW, height=\myImgH]{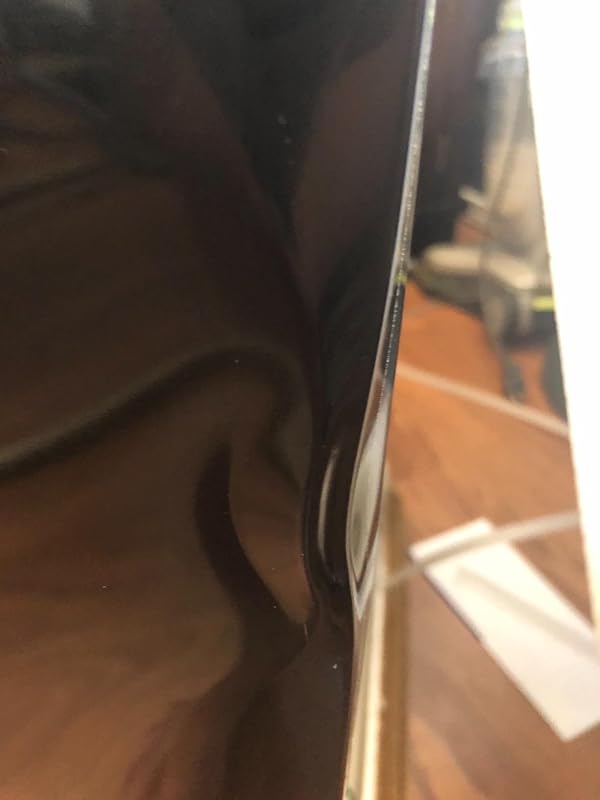} &
\includegraphics[width=\myImgW, height=\myImgH]{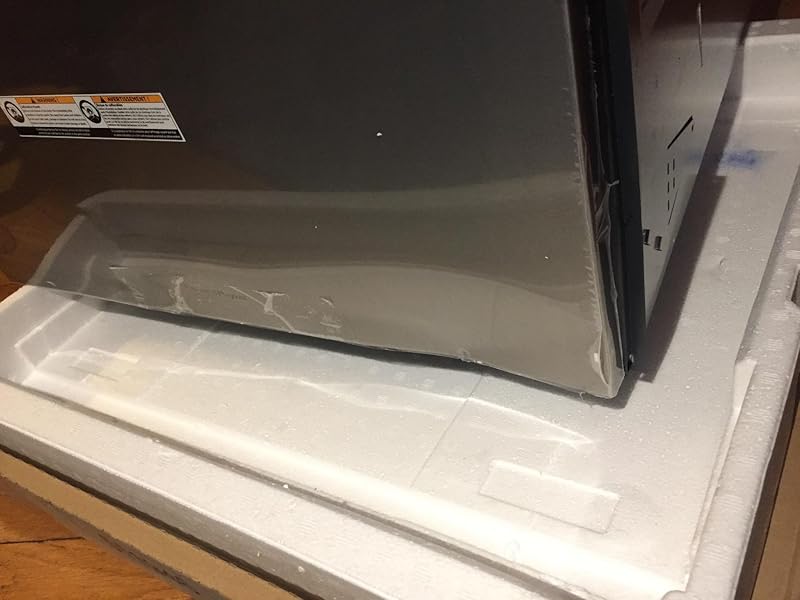} \\

\hline 
&  & & & \\ [\dimexpr-\normalbaselineskip+0.5pt]

\includegraphics[width=\myImgW, height=\myImgH]{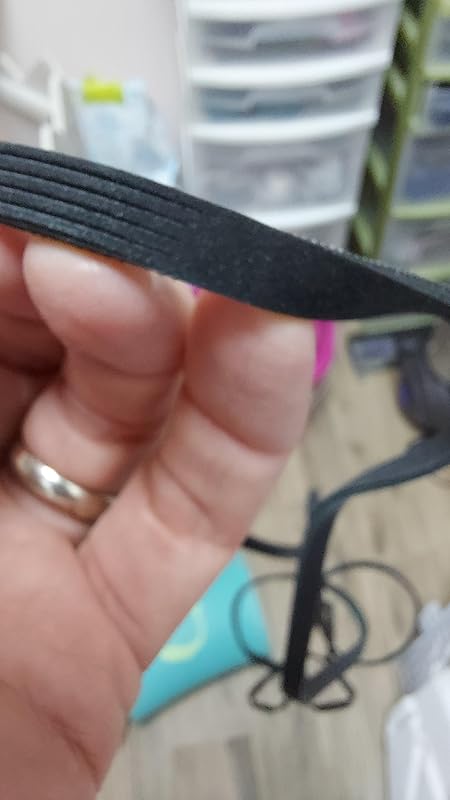} &
\includegraphics[width=\myImgW, height=\myImgH]{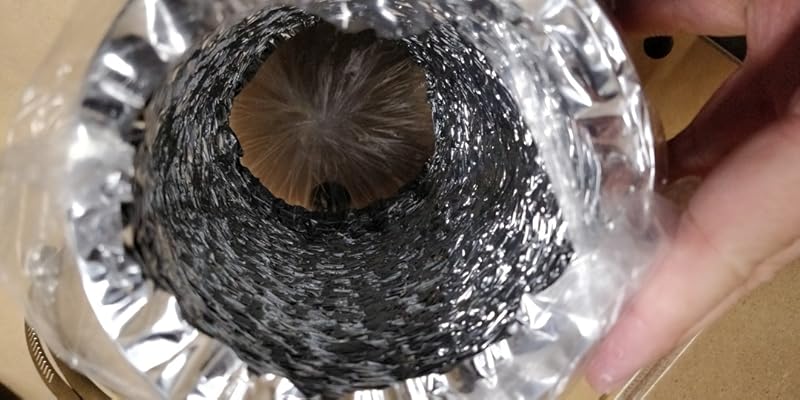} &
\includegraphics[width=\myImgW, height=\myImgH]{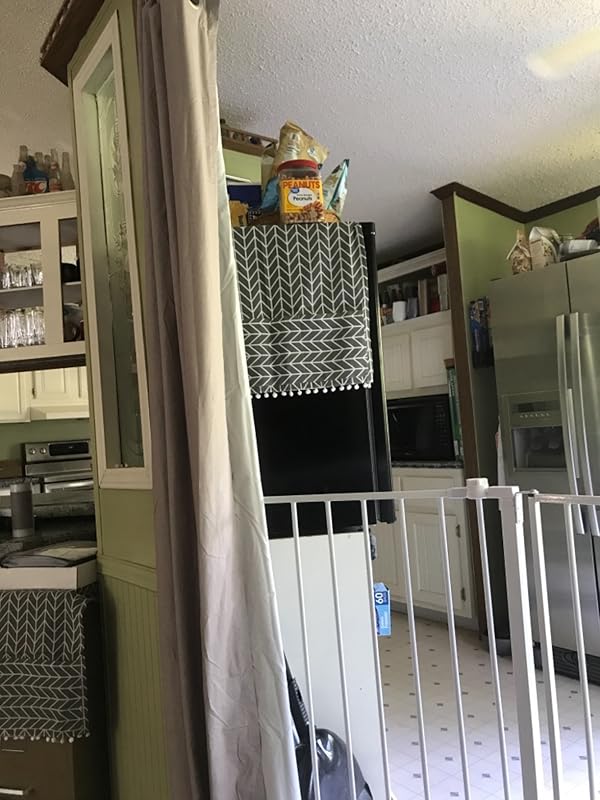} &
\includegraphics[width=\myImgW, height=\myImgH]{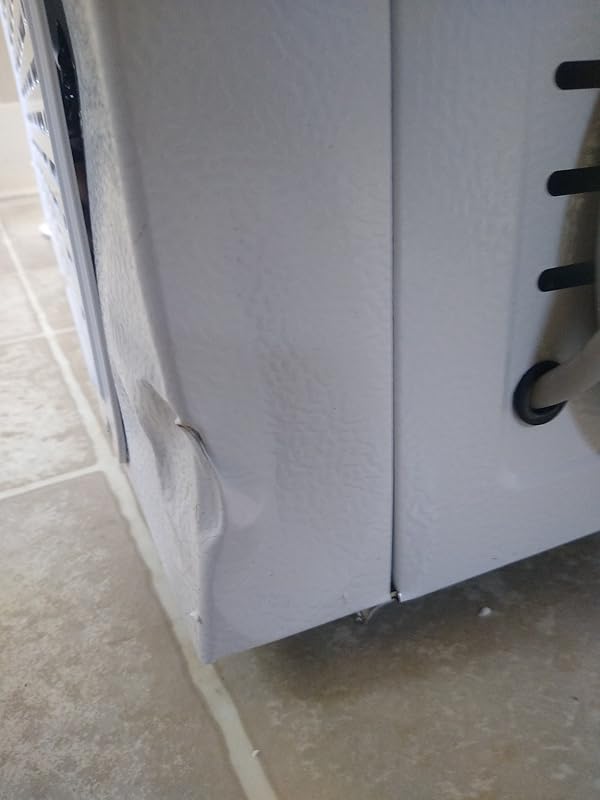} &
\includegraphics[width=\myImgW, height=\myImgH]{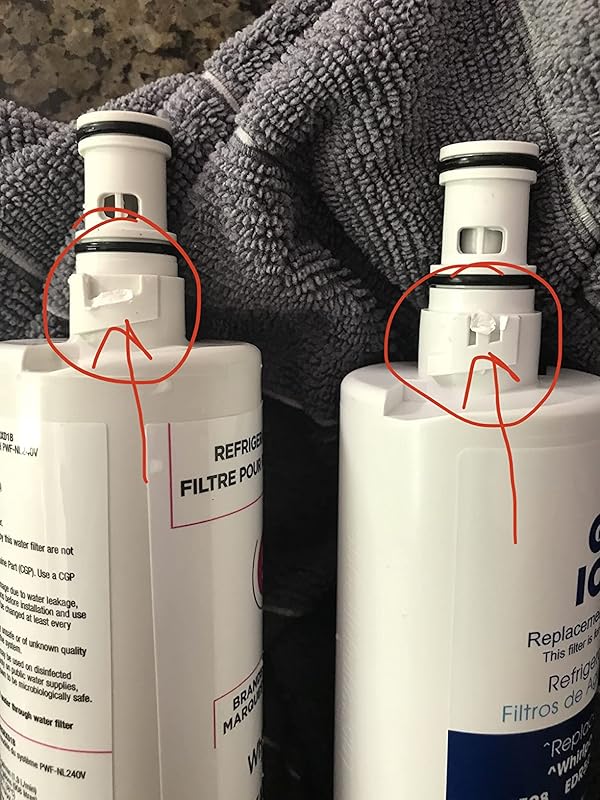} \\

\hline 
&  & & & \\ [\dimexpr-\normalbaselineskip+0.5pt]

\includegraphics[width=\myImgW, height=\myImgH]{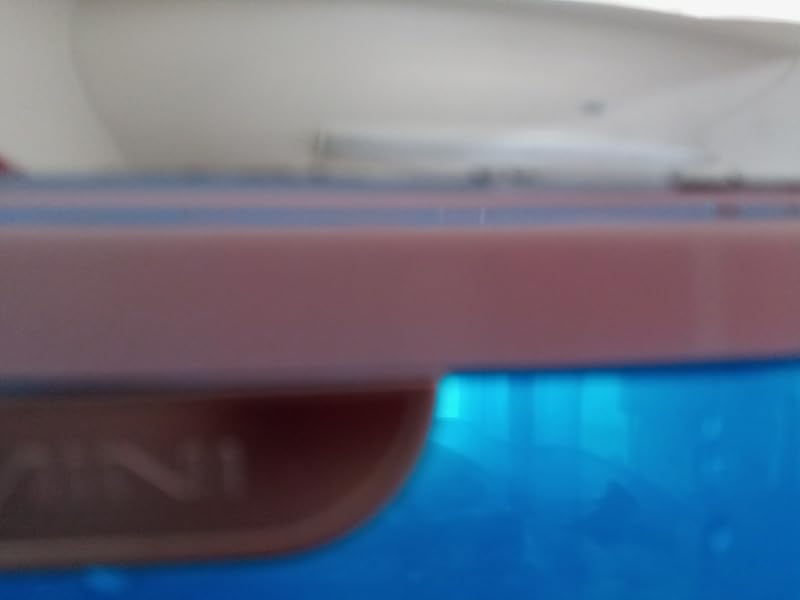} &
\includegraphics[width=\myImgW, height=\myImgH]{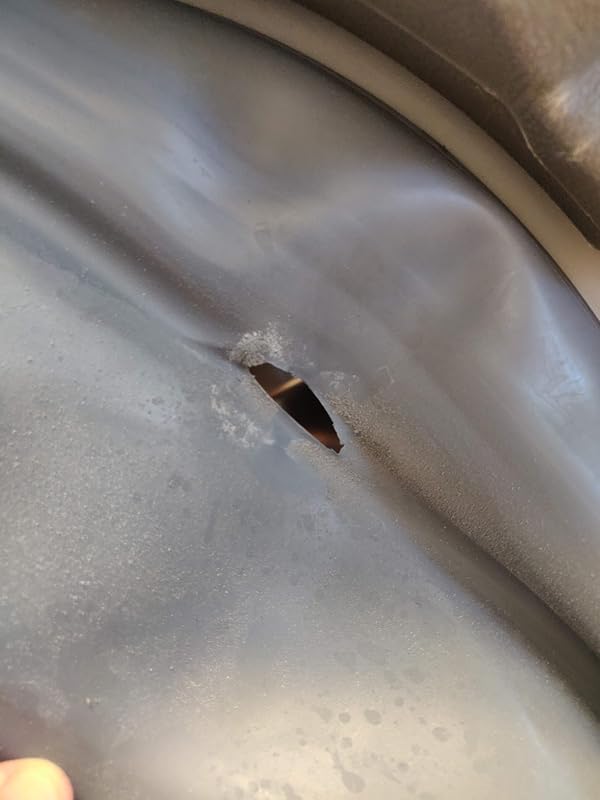} &
\includegraphics[width=\myImgW, height=\myImgH]{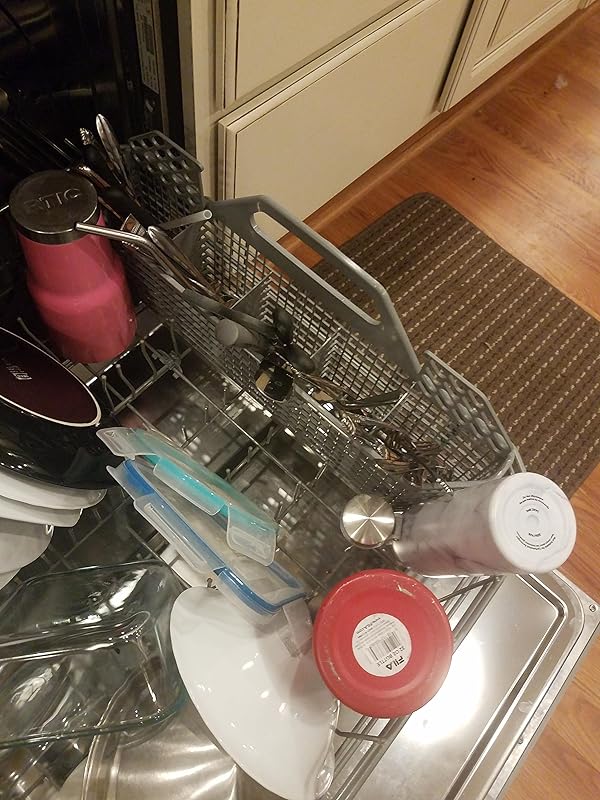} &
\includegraphics[width=\myImgW, height=\myImgH]{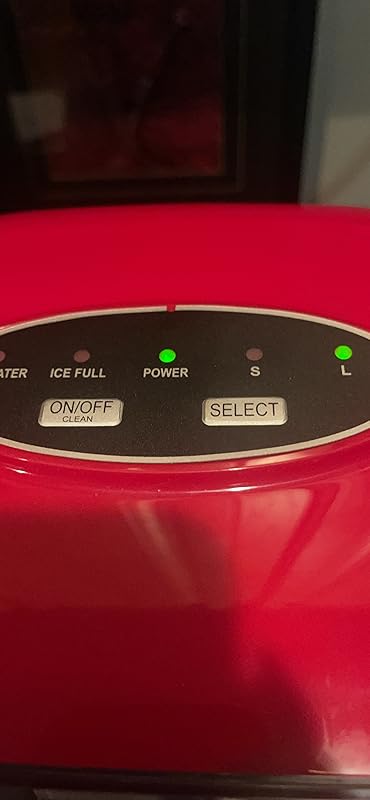} &
\includegraphics[width=\myImgW, height=\myImgH]{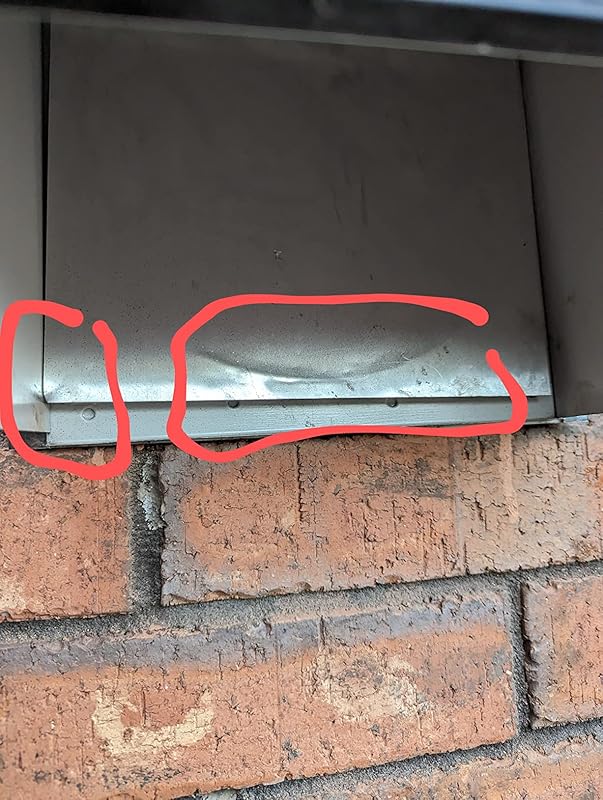} \\ \hline

{\tiny{Degraded quality}} & {\tiny{Zoomed in}} & {\tiny{Target ambiguity}} & {\tiny{Partial View}} & {\tiny{Damaged Product}} \\

\hline
\end{tabular}
\end{adjustbox}
\caption{\small Examples of some challenging product images \cite{amazonDataset}.}
\label{fig:challenge_final}
\end{figure}

The real-world nature of the dataset introduces significant visual noise, including degraded image quality, extreme magnification, target ambiguity, partial view, and damaged product, making image-grounded review generation a highly complex task (see Fig. \ref{fig:challenge_final}). Unlike professionally captured catalog images, user-uploaded images are unconstrained and frequently suffer from severe environmental and compositional variations. Addressing these challenges is crucial for realistic evaluation. It underscores the necessity for a framework that does not merely generate review directly from user images, but rather employs rigorous, step-by-step product identification and visual evidence extraction to accurately ground subjective user sentiment and experience.

\section{Methodology}
\label{sec:method1}

\subsection{Problem Formulation}

Let $\mathcal{D} = \{\mathcal{I}_i, \mathcal{\mathcal{P}}_i, \mathcal{R}_i, \mathcal{T}_i\}^{N}_{i=1}$ be a dataset of $N$ examples, where $\mathcal{I}_i$ is the product image. $\mathcal{\mathcal{P}}_i$ is the product listing text containing the ground-truth product name. $\mathcal{R}_i$ is the product rating in 1 to 5.
 $\mathcal{T}_i = \{{{w}_i}_{1},{{w}_i}_{2},\ldots,{{w}_i}_{t_i}\}$ is the user textual review with $t_i$ length.
Given a product image $\mathcal{I}_i$, the objective is to automatically generate a natural, user-like product review text $\mathcal{T}_i$ that is visually grounded, and product-specific. The task is to learn a generation function \mbox{\( f_\theta: \mathcal{I}_i \rightarrow \mathcal{T}_i \)}.
Instead of treating review generation as a single end-to-end task, we formulate it as a Multi-Agent System, where multiple specialized agents cooperate through a shared memory space. Let the multi-agent system be defined as, $\mathcal{M} = \{\mathcal{A}, \mathcal{S}, \mathcal{I}, \mathcal{\mathcal{P}}, \mathcal{R}, \mathcal{T}\}$, where, $\mathcal{A}$ is the set of agents and $\mathcal{S}$ is the shared memory.  

\begin{figure}[htbp]
    \centering
    
    \includegraphics[width=0.95\linewidth]{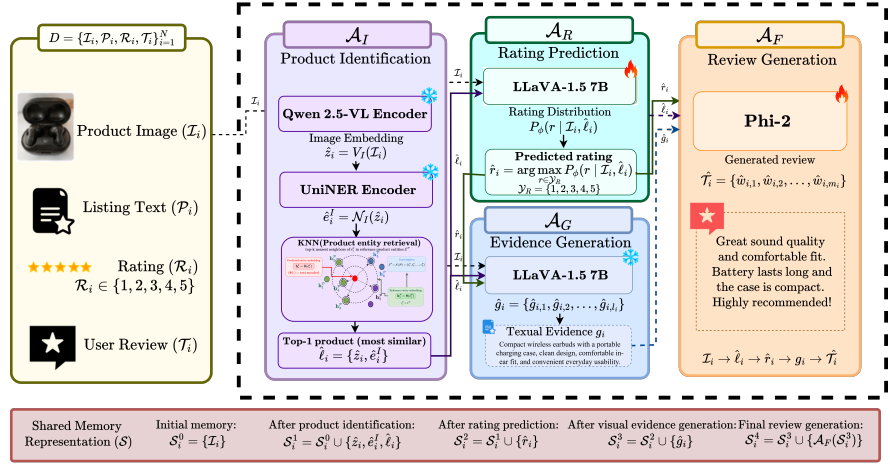}
    \caption{Workflow of the proposed architecture}
    \label{fig:architecture}
   
\end{figure}






\subsection{Solution Architecture}
\noindent
Our objective is to learn an image-grounded review generation function $f_\theta: \mathcal{I}_i \rightarrow \hat{\mathcal{T}}_i$, where $\hat{\mathcal{T}}_i$ is the generated review. Following the formulation in Section 3.1, we implement the task as a sequential multi-agent vision-language system. Each agent reads the current shared memory, performs a specialized operation, and appends its output to the memory for use by downstream agents. 
The agent set is defined as $\mathcal{A}=\{\mathcal{A}_{I},\mathcal{A}_{R},\mathcal{A}_{G},\mathcal{A}_{F}\}$, corresponding to product identification, rating prediction, evidence generation, and final review synthesis, respectively. This decomposition enables product-specific, visually grounded, and sentiment-aware review generation. The overall architecture of the proposed multi-agent vision-language framework is shown in Fig. \ref{fig:architecture}.


\noindent
\textbf{Shared Memory Representation ($\mathcal{S}$):} 
For each input image $\mathcal{I}_i$, the shared memory is initialised as $\mathcal{S}_i^{0}
=\{\mathcal{I}_i\}$. Each agent reads the available information from the shared memory and adds its output to it. After the $k^{th}$ agent, the memory is updated as $\mathcal{S}_i^{k}=\mathcal{S}_i^{k-1}\cup\{o_i^{k}\}$, where $o_i^{k}$ denotes the output of the $k^{th}$ agent. By utilizing shared memory, downstream agents condition their predictions on both the original image and prior reasoning steps. Since the final review generator is a text-only language model, it does not directly process the image. Instead, it receives image-derived intermediate representations, including product identity, predicted rating, and visual evidence generated by the upstream vision-language agents.

\noindent
{\textbf{Product Identification ($\mathcal{A}_{I}$):}} The objective of the product identification agent 
$\mathcal{A}_{I}$ is to perform product-level visual grounding by identifying the product category or product name from the image $\mathcal{I}_i$. This agent comprises two core components: $\mathcal{A}_{I}=\{\mathcal{V}_{I},\mathcal{N}_{I}\}$
where $\mathcal{V}_{I}$ is a vision-language model Qwen2.5-VL-7B \cite{qwen} for visual identification, and $\mathcal{N}_{I}$ is a NER model UniNER-7B \cite{universalner} utilized for entity extraction.
Given $\mathcal{I}_i$, $\mathcal{V}_{I}$ generates $n$ descriptive sequence for product identification $\hat{z}_i =\{\hat{z}_{i,1},\hat{z}_{i,2},\ldots,\hat{z}_{i,n_i}\},$ where $\hat{z}_{i,t}$ is the $t^{th}$ generated token and $n_i$ is the generated sequence length. It can be formulated as: $P_{\omega}\left(\hat{z}_i\mid\mathcal{I}_i\right)=\prod_{t=1}^{n_i}P_{\omega}\left(\hat{z}_{i,t}\mid\hat{z}_{i,<t},\mathcal{I}_i\right)$,
where $\omega$ represents the parameters of $\mathcal{V}_{I}$. 
Although $\hat{z}_i
=\mathcal{V}_{I}(\mathcal{I}_i)$ provides a visual product description, it may contain redundant, generic, or non-informative words. To mitigate this, we applied $\mathcal{N}_{I}$ to extract product-specific entities from the generated identification text $\hat{e}_i^{I}=\mathcal{N}_{I}
(\hat{z}_i).$ Agent $\mathcal{A}_{I}$ maps the input image $\mathcal{I}_i$ into a structured product-grounding representation $\hat{\ell}_i = \{\hat{z}_i, \hat{e}_i^{I}\}$. This representation is then incorporated into the shared memory, updating the initial memory state $\mathcal{S}_i^{0}$ to $\mathcal{S}_i^{1} = \mathcal{S}_i^{0} \cup \{\hat{z}_i, \hat{e}_i^{I}, \hat{\ell}_i\}$.

During the evaluation phase, 
$\mathcal{N}_{I}$ is utilized to extract a set of reference entities from all ground-truth product listings in the training set, denoted as $\mathcal{P}$. This yields a reference entity set $\mathcal{E}^{\mathcal{P}} = \mathcal{N}_{I}(\mathcal{P}) = \{e_{1}^{\mathcal{P}}, e_{2}^{\mathcal{P}}, \ldots, e_{q}^{\mathcal{P}}\}$, where $q$ is the total number of reference entities. To predict the final product category for an unseen image $\mathcal{I}_i$, we embed both the predicted entity $\hat{e}_i^{I}$ and the reference entities $\mathcal{E}^{\mathcal{P}}$ using a text encoder $\Phi(\cdot)$. This results in the embedding $\mathbf{h}_{i}^{I} = \Phi(\hat{e}_{i}^{I})$ for the prediction, and $\mathbf{h}_{j}^{\mathcal{P}} = \Phi(e_{j}^{\mathcal{P}})$ for each reference entity $e_{j}^{\mathcal{P}} \in \mathcal{E}^{\mathcal{P}}$. Finally, we apply the KNN algorithm over these embeddings to assign the most probable product category to $\mathcal{I}_i$.

\noindent
{\textbf{Rating Prediction ($\mathcal{A}_{R}$):}} 
The rating prediction agent $\mathcal{A}_{R}$ utilizes LLaVa-1.5-7b \cite{llava}. It is designed to predict the product rating conditioned on the image $\mathcal{I}_i$ and the product-grounding representation $\hat{\ell}_i$ obtained from $\mathcal{A}_{I}$. $\mathcal{A}_{R}$ parameterized by $\phi$ outputs a probability $P_{\phi}(r \mid \mathcal{I}_i, \hat{\ell}_i)$ for each possible rating $r \in \mathcal{Y}_{R}$, where $\mathcal{Y}_{R}=\{1,2,3,4,5\}$. The final predicted rating, $\hat{r}_i$, is determined by selecting the most probable rating: $\hat{r}_i=\arg\max_{r\in\mathcal{Y}_{R}}P_{\phi}(r\mid\mathcal{I}_i,\hat{\ell}_i)$. This prediction $\hat{r}_i$ functions as an explicit sentiment prior, effectively guiding the downstream visual evidence and review generation agents. Following this prediction step, the shared memory state is updated to include the predicted rating as $\mathcal{S}_i^{2} = \mathcal{S}_i^{1} \cup \{\hat{r}_i\}$.

\noindent
{\textbf{Evidence Generation ($\mathcal{A}_{G}$):}} 
The purpose of the evidence agent $\mathcal{A}_{G}$ is to generate product- and rating-aware evidence. Utilizing the LLaVA-1.5-7B model \cite{llava}, the agent processes the input image $\mathcal{I}_i$, product-grounding representation $\hat{\ell}_i$, and predicted rating $\hat{r}_i$. This process is formulated as:  $\hat{g}_i
=\mathcal{A}_{G}(\mathcal{I}_i,\hat{\ell}_i,\hat{r}_i)$. The resulting evidence is denoted as a sequence of tokens $\hat{g}_i=\{\hat{g}_{i,1},\hat{g}_{i,2},\ldots,\hat{g}_{i,l_i}\},$ where $\hat{g}_{i,t}$ is the $t^{th}$ generated token and $l_i$ is the sequence length.  The generated visual evidence $\hat{g}_i$ acts as an intermediate semantic bridge between visual perception and final review synthesis by encoding product-specific visual cues under the sentiment prior $\hat{r}_i$. The shared memory is updated to incorporate the new evidence: $\mathcal{S}_i^{3}=\mathcal{S}_i^{2}\cup\{\hat{g}_i\}$.

\noindent
{\textbf{Final Review Generation ($\mathcal{A}_{F}$):}} $\mathcal{A}_{F}$ generates the final review by utilizing previous agent outputs, the product-grounding representation $\hat{\ell}_i$, predicted rating $\hat{r}_i$, and evidence $\hat{g}_i$ as $\hat{\mathcal{T}}_i = \mathcal{A}_{F}(\hat{\ell}_i,\hat{r}_i,\hat{g}_i)$. Here, we employed Phi-2 \cite{javaheripi2023phi}. The conditional generation probability is computed as: $P_{\theta}\left(\hat{\mathcal{T}}_i\mid \hat{\ell}_i,\hat{r}_i,\hat{g}_i\right)=\prod_{t=1}^{m_i}P_{\theta}\left({w}_{i,t}\mid {w}_{i<t}, \hat{\ell}_i,\hat{r}_i,\hat{g}_i\right) $
where, $\hat{\mathcal{T}}_i=\{\hat{w}_{i,1},\hat{w}_{i,2},\ldots,\hat{w}_{i,m_i}\}$ is the final generated review, and $\hat{w}_{i,t}$ is the $t^{th}$ token and $m_i$ is the review length.

\noindent
{\textbf{Overall Training Objective:}} The framework is trained modularly by optimising the rating prediction and review generation agents. Let  $\phi$ and $\theta$ denote the trainable parameters corresponding to $\mathcal{A}_{R}$ and $\mathcal{A}_{F}$, respectively. The objective $\mathcal{A}_{R}$ and $\mathcal{A}_{F}$ to minimize $\mathcal{L}_R$ and $\mathcal{L}_F$, respectively. $\mathcal{L}_R$ and $\mathcal{L}_F$ defined as:
\begin{equation}
\scriptsize
\mathcal{L}_{R}= - \sum_{r\in\mathcal{Y}_{R}}[\mathcal{R}_i=r] \log P_{\phi}\left(r\mid{\mathcal{I}}_i,\hat{\ell}_i\right) 
; \mathcal{L}_{F}= - \frac{1}{m_i}\sum_{t=1}^{m_i}\log P_{\theta}\left(w_{i,t}\mid {w}_{i,<t},\hat{\ell}_i,\hat{r}_i,\hat{g}_i\right)
\end{equation}


During training, product listings, ratings, and ground-truth reviews are used as supervision signals. During inference, only the product image ${\mathcal{I}}_i$ is provided, and all intermediate representations and the final review are generated by the proposed framework.
The process will obtain intermediate output $\hat{z}_i=\mathcal{V}_{{I}}({\mathcal{I}}_i)$, $\hat{e}_i^{{I}}=\mathcal{N}_{{I}}(\hat{z}_i)$, $\hat{\ell}_i=\{\hat{z}_i,\hat{e}_i^{{I}}\}$, $\hat{r}_i=\mathcal{A}_{{R}}({\mathcal{I}}_i,\hat{\ell}_i)$, $\hat{g}_i=\mathcal{A}_{{G}}({\mathcal{I}}_i,\hat{\ell}_i,\hat{r}_i)$, and the final generated review $\hat{\mathcal{T}}_i=\mathcal{A}_{{F}}(\hat{\ell}_i,\hat{r}_i,\hat{g}_i)$.

\label{sec:method}

\section{Experiments and Results}
\label{4sec:exp}

All experiments were conducted on a system equipped with an NVIDIA RTX A6000 GPU with 48 GB VRAM and 256 GB RAM. The dataset is partitioned into a training, validation, and test sets with 8:1:1 ratio. $\mathcal{A}_{R}$ and $\mathcal{A}_{F}$ were finetuned using LoRA \cite{lora}, where the LoRA rank and scaling factor are set to 64 and 32, respectively. The framework optimization was performed using the AdamW optimizer \cite{adamw} with learning rate $2 \times 10^{-5}$ for a maximum of 50 epochs. To mitigate overfitting, we adopt an early stopping strategy with a patience of 3 epochs based on validation performance. All results are reported on the test set.

We evaluated performance using accuracy for product identification and rating prediction. For evidence and review generation, we assessed using JudgeLM-7B \cite{judgelm} as LLM-as-Judge ($S_{JLM}$) evaluates generated reviews on a five-point scale, z-score normalized precision BARTScore  ($B_P$), recall BARTScore ($B_R$), F1 BARTScore ($B_{F1}$) \cite{bartscore}, F1 BERTScore ($BS_{F1}$) \cite{bert_score}, MoverScore ($MS$) \cite{moverscore}, Flesch Reading Ease score ($FRES$)\cite{flesch}, CLIPScore ($S_{CLIP}$) \cite{clipscore}, and CoLA-based linguistic acceptability score ($CoLA$) \cite{cola}.


\begin{table}[b]
    \centering
    \caption{Performance comparison of proposed framework with baselines}
    \label{tab:overall1}
    \scriptsize
    \begin{adjustbox}{width=0.9\textwidth}
    \begin{tabular}{l|c|c|c|c|c|c|c|c|c}
    \hline
        \textbf{Model} & $S_{JLM}$ & $B_P$ & $B_R$ & $B_{F1}$ & $BS_{F1}$ & $MS$ & $FRES$ & $S_{CLIP}$ &  $CoLA$ \\ \hline

        ResNet-50V2 \cite{resnet} + LSTM \cite{lstm} & 3.258  & -0.659 & 0.097 & -0.293 & 0.808 & 0.727 & 39.480 & 0.212 & 0.289 \\

        LLaVa-1.5-7B \cite{llava} & 3.575  & -0.372 & -0.183 & -0.380 & 0.825 & 0.760 & 60.244 & 0.209 & 0.289 \\ 
        Qwen2.5-VL-7B \cite{qwen} & 3.688  & 0.016 & 0.302 & 0.291 & 0.824 & 0.756 & 50.634 & 0.215 & 0.405 \\
        SmolVLM-Instruct \cite{smolvlm} & 2.978  & 0.063 & -0.373 & -0.300 & 0.818 & 0.708 & 66.545 & 0.212 & 0.325 \\ 
        
        Ours with LLaVa-1.5-7B  & 3.440  & 0.031 & 0.241 & 0.235 & 0.810 & 0.758 & 62.426 & 0.205 & 0.310 \\ 
        Ours with Qwen2.5-VL-7B  & 3.570  & 0.056 & 0.243 & 0.252 & 0.803 & 0.768 & 64.183 & 0.215 & 0.331 \\ 
        Ours with SmolVLM-Instruct  & 3.430  & 0.115 & 0.238 & 0.316 & 0.830 & 0.763 & 71.658 & 0.212 & 0.319 \\ 

        Ours (Proposed) & \textbf{3.832}  & \textbf{0.141} & \textbf{0.319} & \textbf{0.369} & \textbf{0.838} & \textbf{0.775} & \textbf{72.156} & \textbf{0.224} & \textbf{0.501} \\ 
        \hline
    \end{tabular}
    \end{adjustbox}
\end{table}


\begin{table}[!t]
    \centering
    \caption{\small Class-wise performance of proposed framework with various popular VLMs}
    \label{tab:overall2}
    \begin{adjustbox}{width=0.7\linewidth}
    \begin{tabular}{l|l|c|c|c|c|c|c|c|c|c}
    \hline
    \textbf{Category} & \textbf{Model} & $S_{JLM}$ & $B_P$ & $B_R$ & $B_{F1}$ & $BS_{F1}$ & $MS$ & $FRES$ & $S_{CLIP}$ &  $CoLA$ \\ \hline

    \multirow{8}{*}{\shortstack[l]{Commercial\\Food\\Packaging\\Equipment}}  & Resnet-50V2  + LSTM  & 3.333 & -1.424 & 0.483 & -0.625 & 0.820 & 0.696 & 38.171 & 0.216 & 0.283 \\ 
 & LLaVa-1.5-7B  & 3.500 & -0.610 & 0.506 & -0.030 & 0.847 & 0.741 & 64.804 & 0.219 & 0.324 \\ 
 & Qwen2.5-VL-7B  & 3.333 & 0.068 & 0.540 & 0.479 & 0.826 & 0.759 & 58.402 & 0.194 & 0.414 \\ 
 & SmolVLM-Instruct  & 3.000 & -0.280 & 0.352 & 0.082 & 0.833 & 0.660 & 69.079 & 0.213 & 0.405 \\ 
 & Ours with LLaVa-1.5-7B & 3.763 & 0.080 & 0.320 & 0.311 & 0.808 & 0.732 & 62.627 & 0.201 & 0.315 \\ 
 & Ours with Qwen2.5-VL-7B & 3.690 & 0.112 & 0.320 & 0.335 & 0.803 & 0.778 & 64.661 & 0.222 & 0.316 \\ 
 & Ours with SmolVLM-Instruct & 3.648 & 0.208 & 0.317 & 0.400 & 0.829 & 0.738 & 67.308 & 0.203 & 0.350 \\ 
 & Ours & 4.333 & 0.254 & 0.538 & 0.609 & 0.834 & 0.742 & 71.201 & 0.201 & 0.471 \\ \hline

 \multirow{8}{*}{Humidifier}
 & Resnet-50V2 + LSTM & 3.233 & -0.291 & 0.005 & -0.203 & 0.807 & 0.733 & 35.810 & 0.211 & 0.299 \\ 
 & LLaVa-1.5-7B & 3.400 & -0.326 & -0.420 & -0.565 & 0.830 & 0.744 & 53.690 & 0.228 & 0.276 \\ 
 & Qwen2.5-VL-7B & 3.133 & 0.063 & 0.249 & 0.243 & 0.821 & 0.767 & 40.199 & 0.216 & 0.414 \\ 
 & SmolVLM-Instruct & 2.900 & -0.027 & -0.732 & -0.602 & 0.836 & 0.708 & 75.488 & 0.223 & 0.324 \\ 
 & Ours with LLaVa-1.5-7B & 3.097 & 0.071 & -0.003 & 0.048 & 0.810 & 0.779 & 58.983 & 0.206 & 0.304 \\ 
 & Ours with Qwen2.5-VL-7B & 3.023 & 0.083 & 0.006 & 0.063 & 0.803 & 0.818 & 55.877 & 0.226 & 0.353 \\ 
 & Ours with SmolVLM-Instruct & 2.915 & 0.191 & -0.033 & 0.109 & 0.831 & 0.782 & 68.619 & 0.216 & 0.301 \\ 
 & Ours & 3.733 & 0.223 & 0.251 & 0.358 & 0.827 & 0.766 & 76.000 & 0.216 & 0.464 \\ \hline

 \multirow{8}{*}{\shortstack[l]{Oven \&\\Cooktops}}
 & Resnet-50V2 + LSTM & 3.533 & -0.625 & -0.136 & -0.551 & 0.811 & 0.736 & 35.266 & 0.211 & 0.261 \\ 
 & LLaVa-1.5-7B & 3.533 & -0.306 & -0.571 & -0.671 & 0.836 & 0.759 & 60.213 & 0.231 & 0.296 \\ 
 & Qwen2.5-VL-7B & 3.867 & 0.056 & 0.194 & 0.195 & 0.825 & 0.768 & 50.630 & 0.226 & 0.418 \\ 
 & SmolVLM-Instruct & 2.933 & 0.068 & -0.816 & -0.602 & 0.840 & 0.705 & 65.066 & 0.219 & 0.304 \\ 
 & Ours with LLaVa-1.5-7B & 3.532 & 0.119 & 0.117 & 0.177 & 0.808 & 0.753 & 61.630 & 0.206 & 0.310 \\ 
 & Ours with Qwen2.5-VL-7B & 3.855 & 0.102 & 0.118 & 0.166 & 0.801 & 0.810 & 64.642 & 0.232 & 0.319 \\ 
 & Ours with SmolVLM-Instruct & 2.848 & 0.201 & 0.141 & 0.255 & 0.828 & 0.765 & 70.915 & 0.218 & 0.344 \\ 
 & Ours & 4.183 & 0.243 & 0.188 & 0.322 & 0.830 & 0.770 & 71.296 & 0.219 & 0.521 \\ \hline

 \multirow{8}{*}{\shortstack[l]{Range\\Hood\\Parts}}
 & Resnet-50V2 + LSTM & 3.100 & -0.331 & 0.295 & 0.000 & 0.814 & 0.722 & 42.828 & 0.212 & 0.265 \\ 
 & LLaVa-1.5-7B & 3.733 & -0.356 & 0.250 & -0.053 & 0.839 & 0.756 & 65.580 & 0.231 & 0.288 \\ 
 & Qwen2.5-VL-7B & 3.933 & 0.088 & 0.413 & 0.391 & 0.827 & 0.759 & 52.046 & 0.222 & 0.384 \\ 
 & SmolVLM-Instruct & 3.167 & 0.006 & 0.040 & 0.036 & 0.839 & 0.700 & 61.340 & 0.224 & 0.310 \\ 
 & Ours with LLaVa-1.5-7B & 3.263 & 0.060 & 0.388 & 0.352 & 0.812 & 0.732 & 63.916 & 0.208 & 0.313 \\ 
 & Ours with Qwen2.5-VL-7B & 4.223 & 0.117 & 0.386 & 0.390 & 0.805 & 0.781 & 67.040 & 0.230 & 0.336 \\ 
 & Ours with SmolVLM-Instruct & 3.282 & 0.235 & 0.393 & 0.479 & 0.835 & 0.740 & 72.240 & 0.215 & 0.303 \\ 
 & Ours & 3.808 & 0.200 & 0.415 & 0.472 & 0.832 & 0.779 & 72.932 & 0.218 & 0.533 \\ \hline

 \multirow{8}{*}{\shortstack[l]{Refrigerators,\\Freezers \&\\Ice Makers}}
 & Resnet-50V2 + LSTM & 3.133 & -0.430 & -0.269 & -0.519 & 0.804 & 0.734 & 38.771 & 0.212 & 0.282 \\ 
 & LLaVa-1.5-7B & 3.867 & -0.172 & -0.748 & -0.718 & 0.837 & 0.780 & 57.659 & 0.227 & 0.305 \\ 
 & Qwen2.5-VL-7B & 3.400 & 0.077 & 0.130 & 0.158 & 0.825 & 0.803 & 50.047 & 0.206 & 0.403 \\ 
 & SmolVLM-Instruct & 3.000 & 0.301 & -1.205 & -0.747 & 0.839 & 0.714 & 65.075 & 0.215 & 0.340 \\ 
 & Ours with LLaVa-1.5-7B & 3.663 & 0.079 & 0.046 & 0.093 & 0.809 & 0.761 & 60.840 & 0.202 & 0.308 \\ 
 & Ours with Qwen2.5-VL-7B & 3.222 & 0.077 & 0.054 & 0.098 & 0.803 & 0.819 & 62.984 & 0.217 & 0.343 \\ 
 & Ours with SmolVLM-Instruct & 4.082 & 0.252 & 0.061 & 0.227 & 0.832 & 0.769 & 71.341 & 0.207 & 0.314 \\ 
 & Ours & 3.775 & 0.120 & 0.165 & 0.216 & 0.823 & 0.794 & 73.418 & 0.209 & 0.582 \\ \hline

 \multirow{8}{*}{\shortstack[l]{Washer \&\\Dryer}}
 & Resnet-50V2 + LSTM & 3.567 & -0.098 & -0.173 & -0.207 & 0.811 & 0.755 & 46.557 & 0.210 & 0.288 \\ 
 & LLaVa-1.5-7B & 3.633 & -0.129 & -0.619 & -0.584 & 0.833 & 0.769 & 58.227 & 0.227 & 0.293 \\ 
 & Qwen2.5-VL-7B & 3.533 & 0.063 & 0.168 & 0.178 & 0.820 & 0.789 & 52.001 & 0.209 & 0.397 \\ 
 & SmolVLM-Instruct & 3.333 & 0.270 & -0.729 & -0.390 & 0.837 & 0.732 & 66.313 & 0.218 & 0.339 \\ 
 & Ours with LLaVa-1.5-7B & 3.132 & 0.090 & 0.112 & 0.153 & 0.812 & 0.792 & 63.573 & 0.205 & 0.313 \\ 
 & Ours with Qwen2.5-VL-7B & 3.155 & 0.113 & 0.118 & 0.175 & 0.805 & 0.843 & 64.819 & 0.222 & 0.314 \\ 
 & Ours with SmolVLM-Instruct & 3.548 & 0.231 & 0.080 & 0.227 & 0.829 & 0.793 & 71.228 & 0.212 & 0.305 \\ 
 & Ours & 3.512 & 0.089 & 0.182 & 0.208 & 0.824 & 0.787 & 72.024 & 0.212 & 0.521 \\ \hline

 \multirow{8}{*}{\shortstack[l]{Coffee\\Machine}}
 & Resnet-50V2 + LSTM & 2.967 & -1.315 & 0.020 & -0.916 & 0.800 & 0.687 & 37.930 & 0.215 & 0.286 \\ 
 & LLaVa-1.5-7B & 3.300 & -0.344 & -0.347 & -0.520 & 0.835 & 0.755 & 65.203 & 0.230 & 0.341 \\ 
 & Qwen2.5-VL-7B & 4.333 & 0.081 & 0.263 & 0.267 & 0.823 & 0.774 & 57.347 & 0.212 & 0.413 \\ 
 & SmolVLM-Instruct & 2.533 & 0.143 & -0.477 & -0.278 & 0.839 & 0.697 & 60.664 & 0.222 & 0.325 \\ 
 & Ours with LLaVa-1.5-7B & 3.632 & 0.095 & 0.113 & 0.158 & 0.810 & 0.753 & 65.430 & 0.205 & 0.313 \\ 
 & Ours with Qwen2.5-VL-7B & 3.823 & 0.148 & 0.116 & 0.197 & 0.804 & 0.804 & 69.278 & 0.224 & 0.337 \\ 
 & Ours with SmolVLM-Instruct & 3.682 & 0.193 & 0.116 & 0.229 & 0.828 & 0.756 & 70.857 & 0.213 & 0.323 \\ 
 & Ours & 3.867 & 0.216 & 0.268 & 0.367 & 0.828 & 0.761 & 72.515 & 0.215 & 0.450 \\ \hline

    \end{tabular}

    \end{adjustbox}
    
\end{table}

\begin{table}[t]
    \centering
    \caption{Ablation Study}
    \label{table3:ablation}
    \scriptsize
     \begin{adjustbox}{width=0.6\textwidth}
    \begin{tabular}{l|c|c|c|c|c|c|c|c|c}
    \hline
        \textbf{Models} & $S_{JLM}$ & $B_P$ & $B_R$ & $B_{F1}$ & $BS_{F1}$ & $MS$ & $FRES$ & $S_{CLIP}$ &  $CoLA$ \\ \hline

        Ours & \textbf{3.832}  & \textbf{0.141} & \textbf{0.319} & \textbf{0.369} & \textbf{0.838} & \textbf{0.775} & \textbf{72.156} & \textbf{0.224} & \textbf{0.501} \\ 
        Ours - $\mathcal{A}_G$ & 3.644 & 0.125 & 0.301 & 0.334 & 0.825 & 0.758 & 68.164 & 0.212 & 0.416\\
        Ours - $\mathcal{A}_R$ & 3.487 & 0.116 & 0.295 & 0.286 & 0.819 & 0.734 & 66.456 & 0.208 & 0.357\\
        Ours - $\mathcal{A}_I$ & 3.316 & 0.108 & 0.285 & 0.268 & 0.806 & 0.716 & 63.289 & 0.201 & 0.337\\ \hline
        

    \end{tabular}
    \end{adjustbox}
    
\end{table}

\noindent
\textbf{Comparative Study:} 
Table \ref{tab:overall1} presents a comprehensive performance comparison between the proposed multi-agent framework and established baselines. The evaluation includes traditional architectures (ResNet-50V2 + LSTM) alongside state-of-the-art Vision-Language Models, specifically LLaVA-1.5-7B \cite{llava}, Qwen2.5-VL-7B \cite{qwen}, and SmolVLM-Instruct \cite{smolvlm}. Furthermore, we assess several variants of our framework in which the constituent agents are uniformly instantiated with these respective VLMs. The proposed framework consistently outperforms all baseline models across all evaluation metrics. It demonstrates superior semantic fidelity and information recall, achieving normalized scores for BARTScore ($B_{F1}$: 0.369), BERTScore ($BS_{F1}$: 0.838), and MoverScore ($MS$: 0.775). Crucially, this semantic accuracy does not come at the expense of linguistic integrity. The generated text remains highly readable and grammatically sound, yielding Flesch Reading Ease ($FRES$: 72.156) and the Corpus of Linguistic Acceptability ($CoLA$: 0.501). Finally, the framework establishes its qualitative, human-aligned superiority by attaining JudgeLM score ($S_{JLM}$: 3.832). Overall, these findings suggest that the proposed multi-agent architecture successfully generates the coherent, sentiment-driven reviews that traditional end-to-end generation methods inherently lack.

\noindent
\textbf{Class-wise Analysis:} 
Table \ref{tab:overall2} extends the overall performance evaluation by providing a granular, class-wise breakdown across seven distinct product categories, effectively demonstrating the domain-agnostic robustness of the proposed framework. While traditional baselines like ResNet-50V2 + LSTM exhibit severe instability and substantial semantic degradation in specific domains—evidenced by heavily negative normalized $B_P$ values in classes such as ``Commercial Food Packaging Equipment'', the proposed framework maintains consistently high semantic fidelity regardless of the product category. Furthermore, the framework successfully mitigates the linguistic variance frequently observed in standalone VLMs. For instance, while base VLMs show significant fluctuations in $FRES$ across different categories, the proposed architecture acts as a stabilizing constraint, consistently yielding highly readable text alongside $CoLA$. 
The results indicate that the multi-agent formulation can improve several aspects of review generation, although the improvement depends on the underlying VLM and product category.

\begin{table}[b]
    \centering
    \caption{Performance of KNN in product identification phase with various VLMs}
    \scriptsize
    \begin{adjustbox}{width=0.6\textwidth}
    \label{tab1:identification_score}
    \begin{tabularx}{0.6\textwidth}{l|>{\centering\arraybackslash}X|>{\centering\arraybackslash}X}
        \hline
        \textbf{Model} & $Accuracy_{d_{cos}}$ & $Accuracy_{d_{euc}}$ \\
        \hline
        Qwen2.5-VL-7B \cite{qwen}         & \textbf{0.7510} & \textbf{0.7552} \\
        LLaVa-1.5-7B \cite{llava}        & 0.6801 & 0.6815 \\
        SmolVLM-Instruct \cite{smolvlm}     & 0.7455 & 0.7455 \\
        Llama-Vision \cite{llama} & 0.6676 & 0.6704 \\
        Ministral3 \cite{ministral}   & 0.7079 & 0.7079 \\
        \hline
        \multicolumn{3}{r}{\tiny $Accuracy_{d_{cos}}$: Accuracy with cosine distance as distance function,}\\ 
        \multicolumn{3}{r}{\tiny $Accuracy_{d_{euc}}$: Accuracy with euclidean distance as distance function} \\
    \end{tabularx}
    \end{adjustbox}
\end{table}

\noindent
\textbf{Ablation Study:}
Table \ref{table3:ablation} presents a comprehensive ablation study conducted to validate the individual contributions of the constituent components within the proposed framework: the product identification agent ($\mathcal{A}_I$), the rating prediction agent ($\mathcal{A}_R$), and the evidence generation agent ($\mathcal{A}_G$). By systematically omitting each agent, we observe their impacts on the overall architecture. The fully proposed framework achieves the highest performance across all evaluation metrics, demonstrating the complementary contribution of the individual agents.
The most severe performance degradation occurs upon the removal of $\mathcal{A}_I$, which causes $S_{JLM}$ to fall from 3.832 to 3.316 alongside a decline in $CoLA$, and $FRES$. This indicates that accurate early product identification serves as a vital anchor for generating coherent and accessible text downstream. Furthermore, the exclusion of the $\mathcal{A}_R$ reveals its critical role in multimodal grounding and semantic fidelity, evidenced by significant decreases in $S_{CLIP}$ dropping to 0.208 and $B_{F1}$ dropping to 0.286. Finally, while omitting $\mathcal{A}_G$ results in a slightly less severe decline relative to the other modules, its absence still causes a definitive drop in $MS$ and $BS_{F1}$. Overall, the performance degradation observed across all ablated variants empirically validates that the complete, integrated pipeline of product identification, rating prediction, and evidence generation is beneficial.
\noindent
\textbf{Analysis of Agents:}
For product identification, we evaluate five vision-language models and assess category-level correctness, as presented in Table \ref{tab1:identification_score}. The generated product names are compared with UniNER-7B-extracted entities from Amazon listings. We employed sentence-transformers/all-MiniLM-L6-v2 \cite{sentencebert} as text encoder to get embeddings from entities, and used cosine distance or Euclidean distance as the distance function in the KNN algorithm. The analysis reveals Qwen outperformed the other VLMs while SmolVLM performed competitively. The Euclidean distance introduces marginal fractional increases for VLMs. However, their performance remains completely stable across both evaluation metrics.



\begin{table}[t]
    \centering
    \caption{Performance of different models in rating prediction}
    \scriptsize
    \begin{adjustbox}{width=0.6\textwidth}
    
    \begin{tabular}{l|c|c|c|c|c|c}
        \hline
        \multirow{2}{*}{\textbf{Models}} & \textbf{Overall} & \multicolumn{5}{c}{\textbf{Rating wise Accuracy}} \\ \cline{3-7}
                                                                  & \textbf{Accuracy}  & \textbf{1} & \textbf{2} & \textbf{3} & \textbf{4} & \textbf{5} \\ \hline
        
        ResNet50v2 \cite{resnet} &  31.93 & 70.00 & 15.00 & 8.33 & 26.66 & 38.57 \\
        ConvNeXt-B \cite{convnet} &  31.61 & 38.33 & 15.00 & 16.66 & 31.66 & 52.86 \\
        RegNet-Y-128GF \cite{regnet} & 31.29 & 38.33 & 18.33 & 10.00& 23.33 & 61.43  \\
        CvT \cite{cvt} &  30.32 & 28.33 & 21.67 & 0.00 & 5.00 & 87.14 \\
        MaxViT-T \cite{maxvit} &  30.00  & 28.33 & 25.00 & 6.67 & 30.00 & 55.71\\
        DINOv3 \cite{dinov3} & 29.67 & 46.67 & 6.66 & 8.33 & 13.33 & 67.14\\
        ViT-B \cite{vit} &  29.35 & 43.33 & 15.00 & 11.67 & 23.33 & 50.00\\
        LeViT \cite{levit} & 29.35 & 41.67 & 41.67 & 0.00 & 0.00 & 58.57  \\
        SwinV2-B \cite{swinv2}  & 29.03 & 53.33 & 8.33 & 0.00 &6.67 & 70.00 \\
        EfficientNetV2-L \cite{efficientnetv2} & 27.41 & 28.33 & 21.67 & \textbf{20.00} & 28.33 & 37.14 \\
        SqueezeNet \cite{squeezenet} &28.70 & 26.67 & 11.67 & 3.33 & 35.00 & 61.42\\
       
        EfficientNet-B7 \cite{efficientnetv2}  & 28.38 & 23.33 & 33.33 & 13.33 & 30.00 & 40.00 \\
        ResNet152 \cite{resnet} &  28.06 & 38.33 & 3.33 & 16.67 & 25.00 & 52.86\\
        GoogleNet \cite{googlenet} & 27.09 & 20.00 & 28.33 & 6.67 & 26.67 & 50.00 \\
        Wide-ResNet101 \cite{wide} &  26.45 & 48.33 & 15.00 &6.66 & 26.67 & 34.28 \\
        ShuffleNetV2-X2-0 \cite{shufflenetv2}  & 26.12 & 60.00 & 15.00 & 0.00 & 21.67 & 32.86 \\
        
        VGG19 \cite{vgg} &  24.83  & 53.33 & 0.00 & 0.00 & 1.67 & 62.85\\
        AlexNet \cite{alex} &  23.54 & 28.33 & \textbf{43.33} & 5.00 & 3.33 & 35.71 \\
        Ministral3 \cite{ministral} & 28.06 & 8.33 & 13.33 & 6.67 & 5.00 & \textbf{95.71} \\
        DINOv2 \cite{dinov2} &  22.90 & 11.67 & 0.00 & 0.00 & 0.00 & 91.42 \\
        SmolVLM-Instruct \cite{smolvlm}  & 25.81  & \textbf{80.00} & 6.67 & 8.33 & 6.67 & 12.86 \\
        Llama Vision \cite{llama} & 19.35 & 18.33 & 5.00 & 8.33 & 55.00 & 11.43 \\
        Qwen2.5-VL-7B \cite{qwen}  & 18.71 & 6.67 & 1.67 & 0.00 & \textbf{88.33} & 0.00 \\
        Llava-1.5-7B \cite{llava}     & \textbf{32.90} & 61.67 & 18.33 & 11.67 & 5.00 & 62.86 \\
        \hline
    \end{tabular}
    \end{adjustbox}
    
    \label{tab:rating_metrics}
\end{table}

\begin{figure}[!h]
    \centering
    \scriptsize
    
    \begin{adjustbox}{width=0.98\textwidth}
    
    \begin{tabular}{c|lp{12cm}}
    
    \hline 

    \multicolumn{3}{c}{\cellcolor{green1} \textbf{(a) Relevant review generation}}\\ 
    \hline
    \hline

     & \\ [\dimexpr-\normalbaselineskip+1.5pt]
     \multirow{9}{*}{\includegraphics[width=0.1\linewidth]{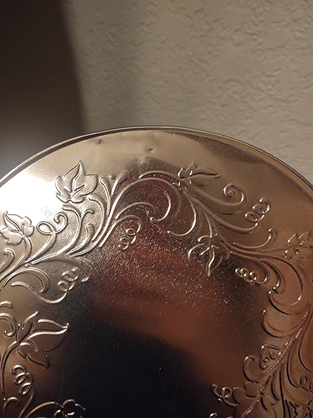}} & Actual Product & Copper, Ivy Embossed Range/Stove Burner Covers \\
    & Actual Rating & 1 \\
    & Actual Review & Cheap material! It's like tin foil. Not to mention it came damaged and doesnt even fit properly. Waste of money! \\
    & Predicted Product & Stove Burner Covers. \\
    & Predicted Rating & 1 \\
    & Generated Review & I recently received a set of these and found the material incredibly cheap, it feels just like tin foil. Although the design looked nice online, several pieces arrived completely damaged. To make matters worse, they don't even fit properly. A total waste of money. \\
    \hline

     & \\ [\dimexpr-\normalbaselineskip+1.5pt]
     \multirow{10}{*}{\includegraphics[width=0.1\linewidth]{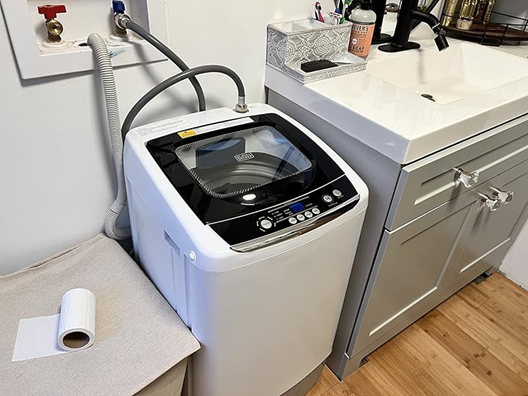}} & Actual Product & BLACK+DECKER Small Portable Washer, Washing Machine for Household Use, Portable Washer \\
    & Actual Rating & 5 \\
    & Actual Review & This washing machine looks SUPER small when you get it out of the box, but it surprisingly washes a lot more than you think (of clothing anyway). it will fit about 4 thick towels. Today I washed king size sheet set and about 10 items of clothes and it did it no problem.  Highly recommend! \\
    & Predicted Product & This is a portable washing machine. \\
    & Predicted Rating & 5 \\
    & Generated Review & The water in my tank is slightly warm, it works very well for the type of clothes we wear and saves money and plastic usage in plastic bags for detergent. I keep the unit empty or on about a full bottle of water and detergent. \\
    \hline


    \multicolumn{3}{c}{\cellcolor{red1} \textbf{(b) Review generation due to rating misprediction}}\\ 
    \hline
    \hline

       & \\ [\dimexpr-\normalbaselineskip+1.5pt]
     \multirow{11}{*}{\includegraphics[width=0.1\linewidth]{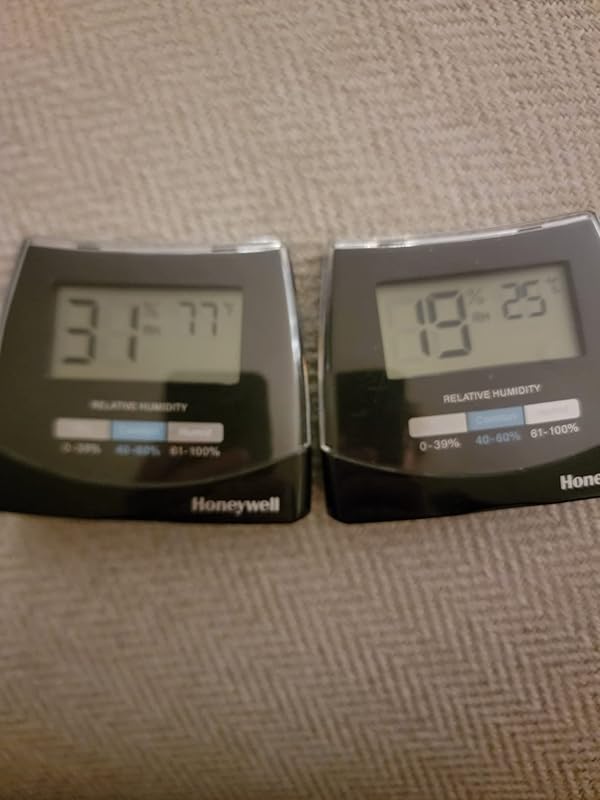}} & Actual Product & Honeywell HHM10B Humidity Monitor \\
    & Actual Rating & 1 \\
    & Actual Review & I make bath products, which can only be made in the most ideal humidity. If the humidity is off, my product doesn't turn out. This happened recently and I was confused as to why. I decided to change the batteries in these as I noticed that they gave me two completely different readings. ... \\
    & Predicted Product & Honeywell digital thermometer and hygrometer. \\
    & Predicted Rating & 5 \\
    & Generated Review & Thermometer display was extremely large and easy to read (wanted it even bigger), battery was able to last for a day without charging and hygrometer readings were also quite good! These temperature and humidity measurements are very accurate and have allowed me to improve the indoor environment in my home. ... \\
    \hline

     & \\ [\dimexpr-\normalbaselineskip+1.5pt]
     \multirow{10}{*}{\includegraphics[width=0.1\linewidth]{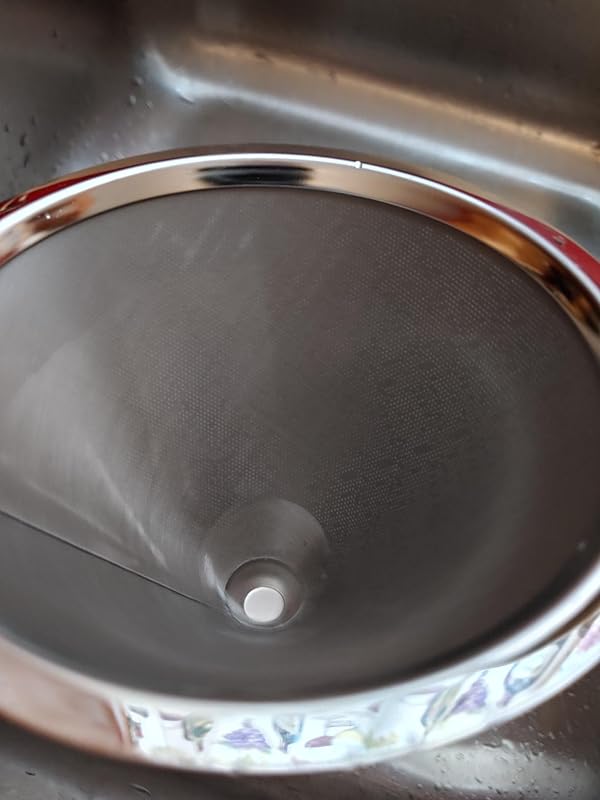}} & Actual Product & Pour Over Coffee Dripper by Soulake - Reusable Stainless Steel Mesh Filter With Drip Stand \\
    & Actual Rating & 2 \\
    & Actual Review & Well, I wanted to like this one. I purchased this to do duel duties for me. I drink tons of coffee and tea and wanted something that could be used for both. So this was my first purchase of a larger stainer that I could use in both manners, or so I thought. 3 major issues for me: ...\\
    & Predicted Product & Metal strainer \\
    & Predicted Rating & 5 \\
    & Generated Review & This is a sturdy metal strainer that drains and stores food easily and well. The strainer has holes that are wide enough to accommodate even large eggplants. The strainer is heavy enough to hold its shape and is easily stack-able, with other pieces sliding into it ...\\
    \hline

    \multicolumn{3}{c}{\cellcolor{red1}\textbf{(c) Review generation due to error propagation}}\\ 
    \hline
    \hline

    & \\ [\dimexpr-\normalbaselineskip+1.5pt]
     \multirow{13}{*}{\includegraphics[width=0.1\linewidth]{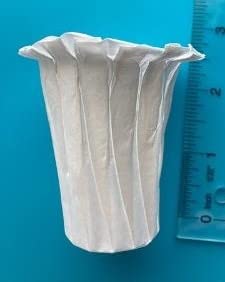}} & Actual Product & BRBHOM Disposable Filters Paper K Carafe Filter Cups K Carafe Pods Compatible Paper \\
    & Actual Rating & 1 \\
    & Actual Review & Then I thoroughly wet the filter and put it back into the holder and tried again.  Same thing --- the filter paper is much too thick to work well.  It isn't made of that filmy almost see through pressed paper coffee filters always seem to be made from.  This was like trying to run water through a piece of notebook paper! ...  \\
    & Predicted Product & sanitary napkin \\
    & Predicted Rating & 5 \\
    & Generated Review & These are great to use in the work environment that I spend my days in. They are disposable with an easy clip for easy access in a drawer or cup holder. I didn\'t notice a significant reduction in odor from my vaginal area at first. It was still there and it was really annoying, but when I put these on, the odor would not come back. I haven\'t stopped wearing them, and my coworkers haven\'t even noticed I\'ve moved over to these from the regular ones for obvious reasons. ... \\
    \hline

     & \\ [\dimexpr-\normalbaselineskip+1.5pt]
    \multirow{10}{*}{\includegraphics[width=0.1\linewidth]{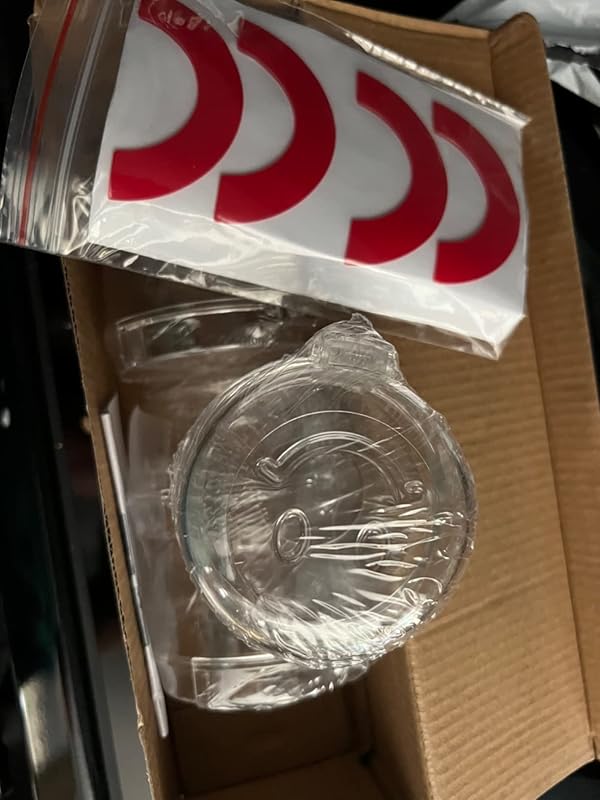}} & Actual Product & Stove Knob Covers(5 Pack) ACM Adhesive 1.26 in Diameter Inner Ring Child Proof  \\
    & Actual Rating & 4 \\
    & Actual Review & They are super easy to install you just take your knob off put these on with the lid flipped up and then put your knob back on. Overall to have these is very handy if you need them, The adhesion works great they aren’t a superheat resistant they may discolor overtime ...  \\
    & Predicted Product & Plastic container  \\
    & Predicted Rating & 1 \\
    & Generated Review & This was a terrible product.  It smelled awful and was slimy when I opened it.  I ordered just two and they have both gotten mold in them.  My recommendation: don’t buy it.  This looks like a cheap product made to look more expensive than it really is.  This is not worth your money. \\
    \hline

\end{tabular}
    \end{adjustbox}
\caption{\small Qualitative analysis of the proposed framework}      
    \label{tab:qualitative_results1}

\end{figure}

\begin{figure}[t]
\centering

\begin{adjustbox}{width=0.8\textwidth}
\begin{tabular}{ C{3.2cm}|C{2.8cm}|C{2.8cm}|C{2.8cm}|C{2.8cm}|C{2.8cm}}
\hline

\multicolumn{6}{c}{\cellcolor{green1} \textbf{(a) Correct Identification}} \\ \hline \hline
& & & & & \\ [\dimexpr-\normalbaselineskip+0.5pt]
 
\textbf{Original Image} & 
\includegraphics[width=2.8cm, height=2.8cm]{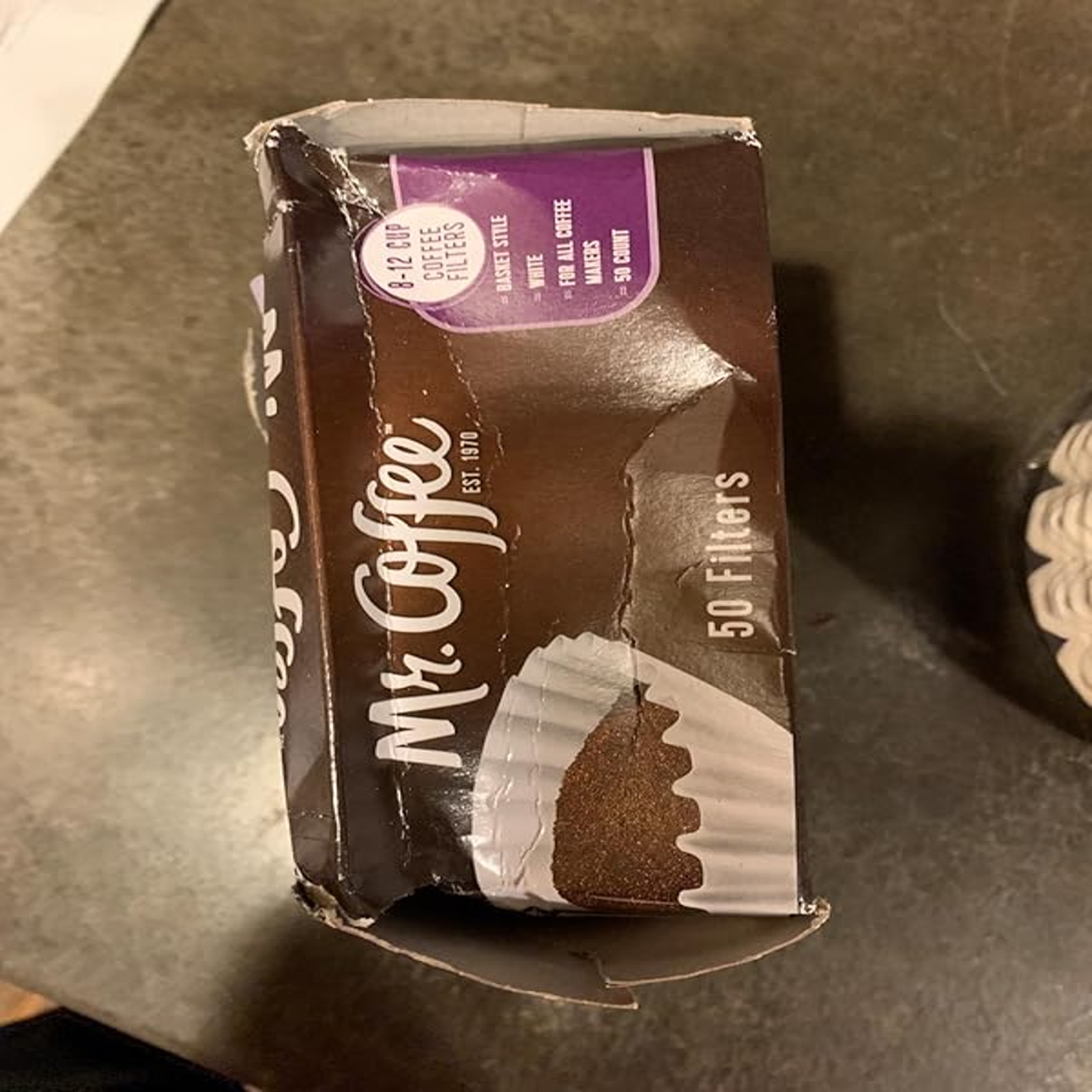} & 
\includegraphics[width=2.8cm, height=2.8cm]{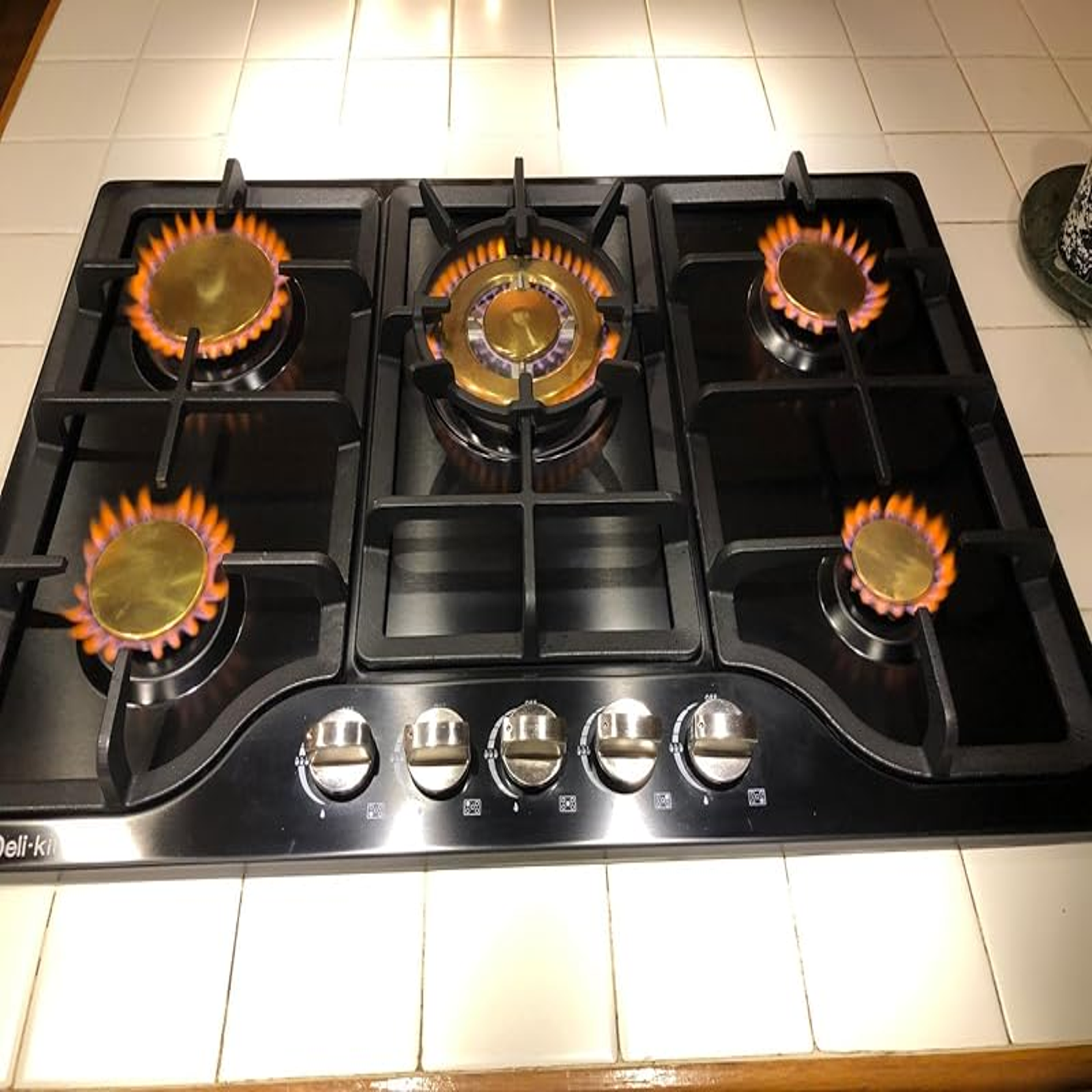} & 
\includegraphics[width=2.8cm, height=2.8cm]{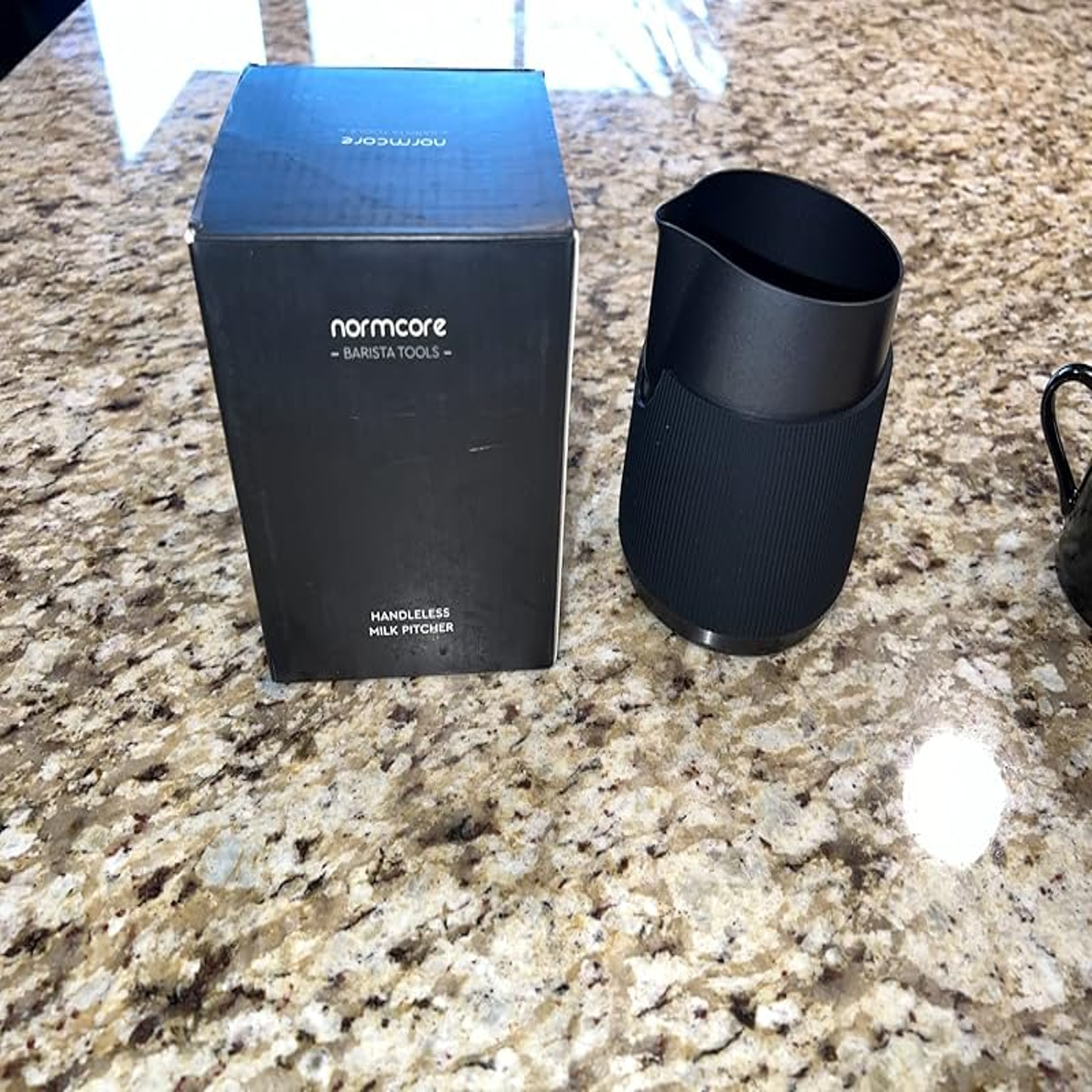} & 
\includegraphics[width=2.8cm, height=2.8cm]{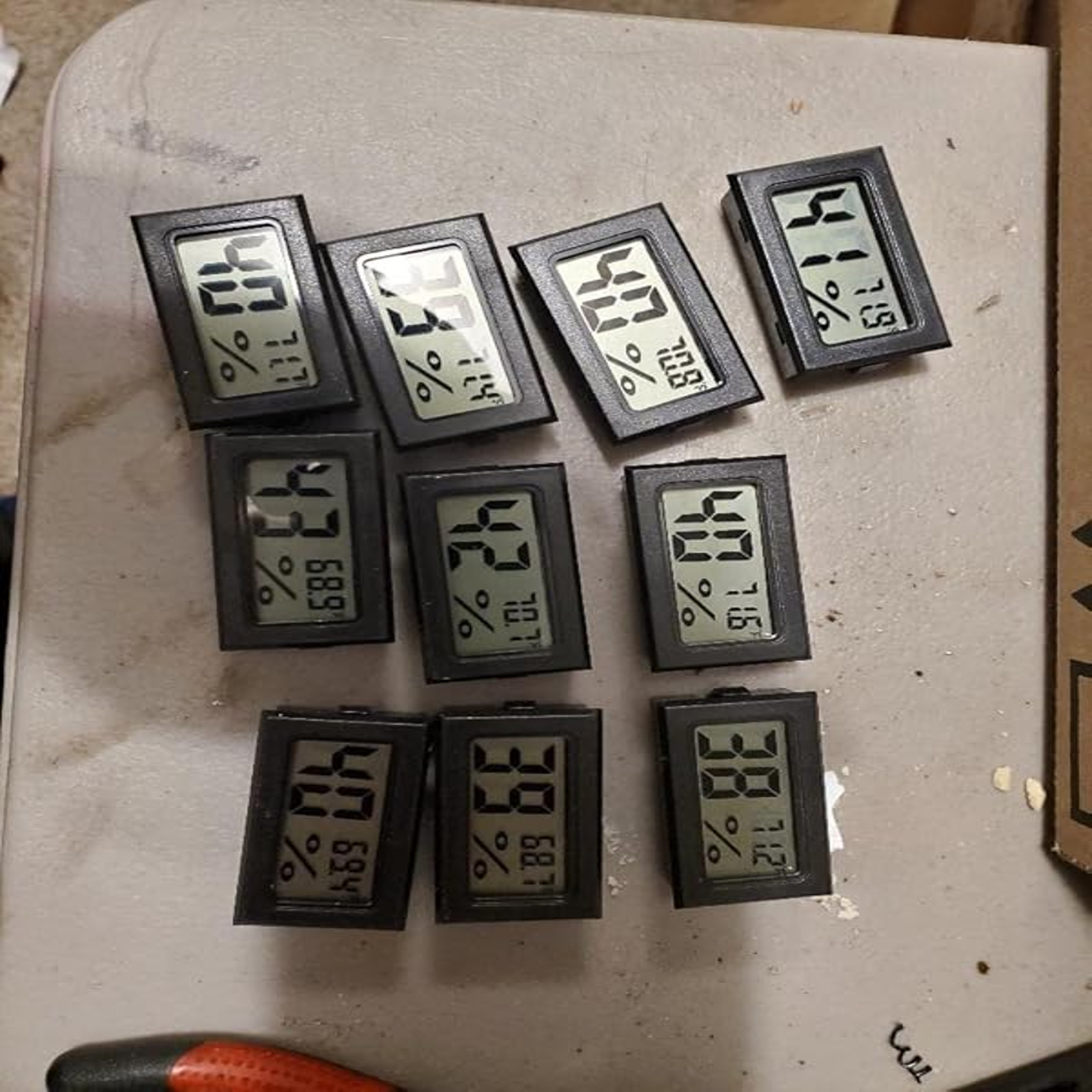} & 
\includegraphics[width=2.8cm, height=2.8cm]{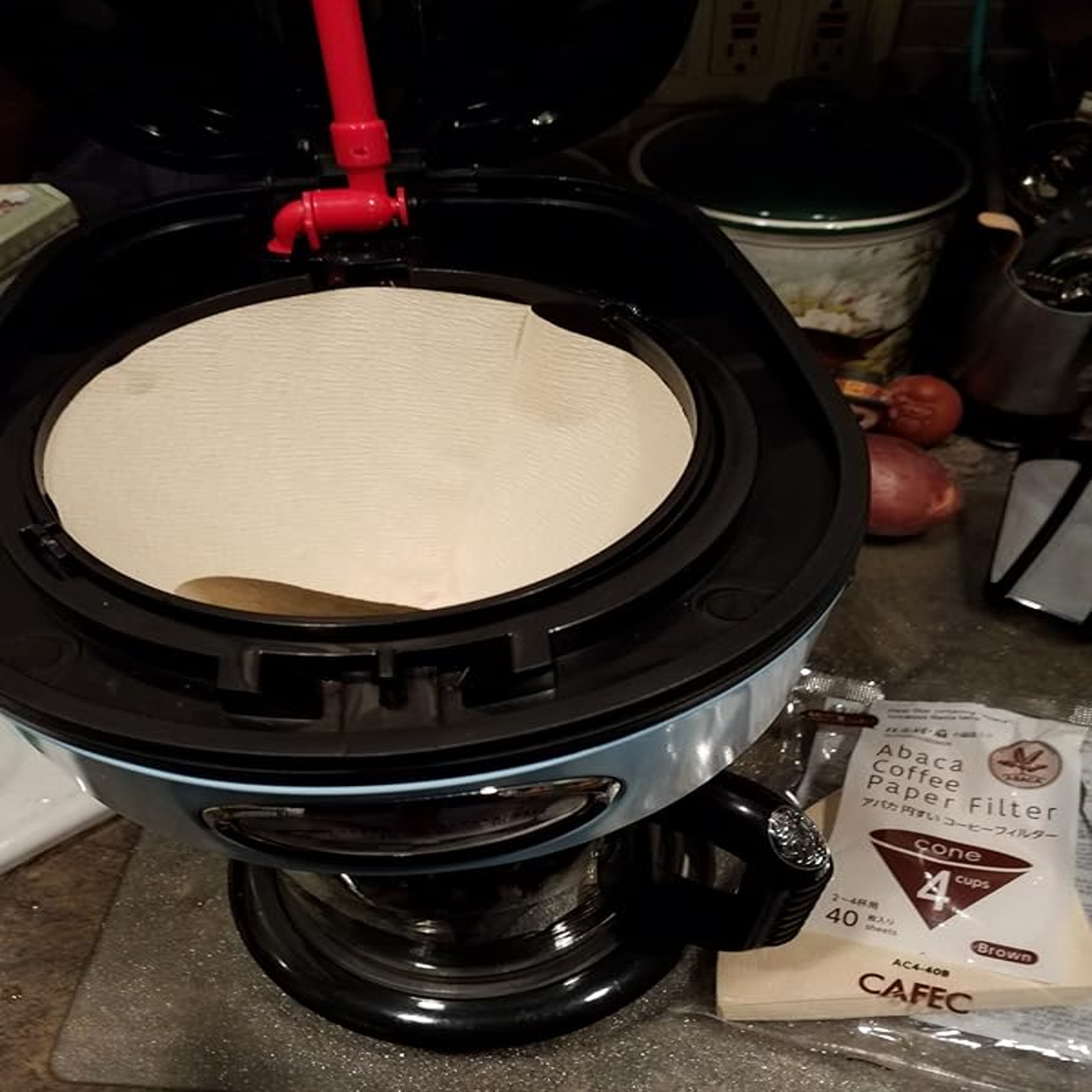} \\ \hline

\multirow{2}{*}{\textbf{Actual Product}} & \multirow{2}{*}{Coffee Filters} & \multirow{2}{*}{Stove} & \multirow{2}{*}{Milk Pitcher} & Small Hygrometer & \multirow{2}{*}{Paper Filter }\\ 

& & & & Thermometer & \\ \hline

 & & & & & \\ [\dimexpr-\normalbaselineskip+0.5pt]
\textbf{Eigen CAM} & 
\includegraphics[width=2.8cm, height=2.8cm]{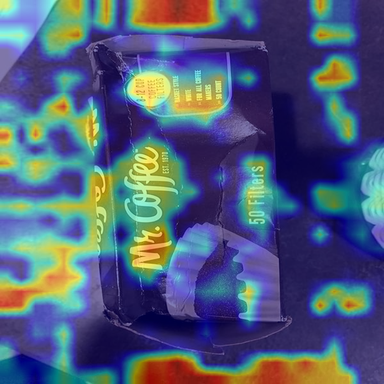} & 
\includegraphics[width=2.8cm, height=2.8cm]{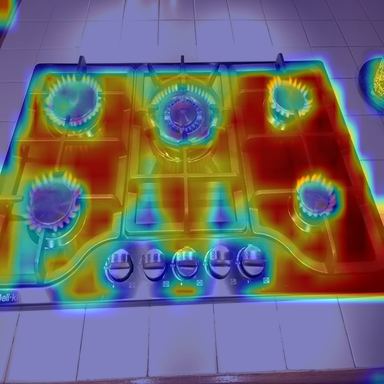} & 
\includegraphics[width=2.8cm, height=2.8cm]{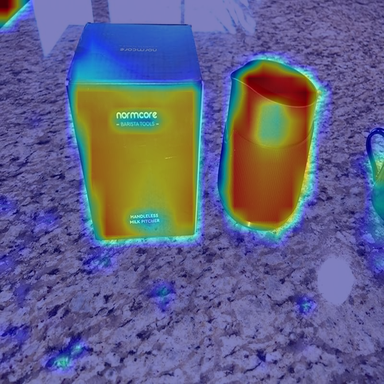} & 
\includegraphics[width=2.8cm, height=2.8cm]{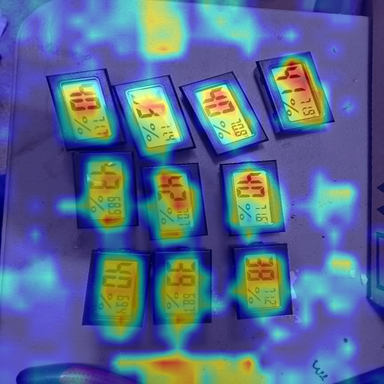} & 
\includegraphics[width=2.8cm, height=2.8cm]{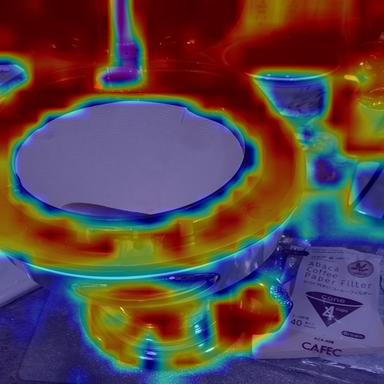} \\ \hline

\multirow{2}{*}{\textbf{Predicted Product}} & \multirow{2}{*}{Coffee filters} & \multirow{2}{*}{Cooktop} & \multirow{2}{*}{Milk pitcher} & \multirow{2}{*}{Thermometer} & \multirow{2}{*}{Coffee filter} \\ 
& & & & & \\

\hline 
& & & & & \\ [\dimexpr-\normalbaselineskip+0.5pt]

\multicolumn{6}{c}{\cellcolor{red1}\textbf{(b) Misidentification}} \\ \hline \hline
\textbf{Original Image} & 
\includegraphics[width=2.8cm, height=2.8cm]{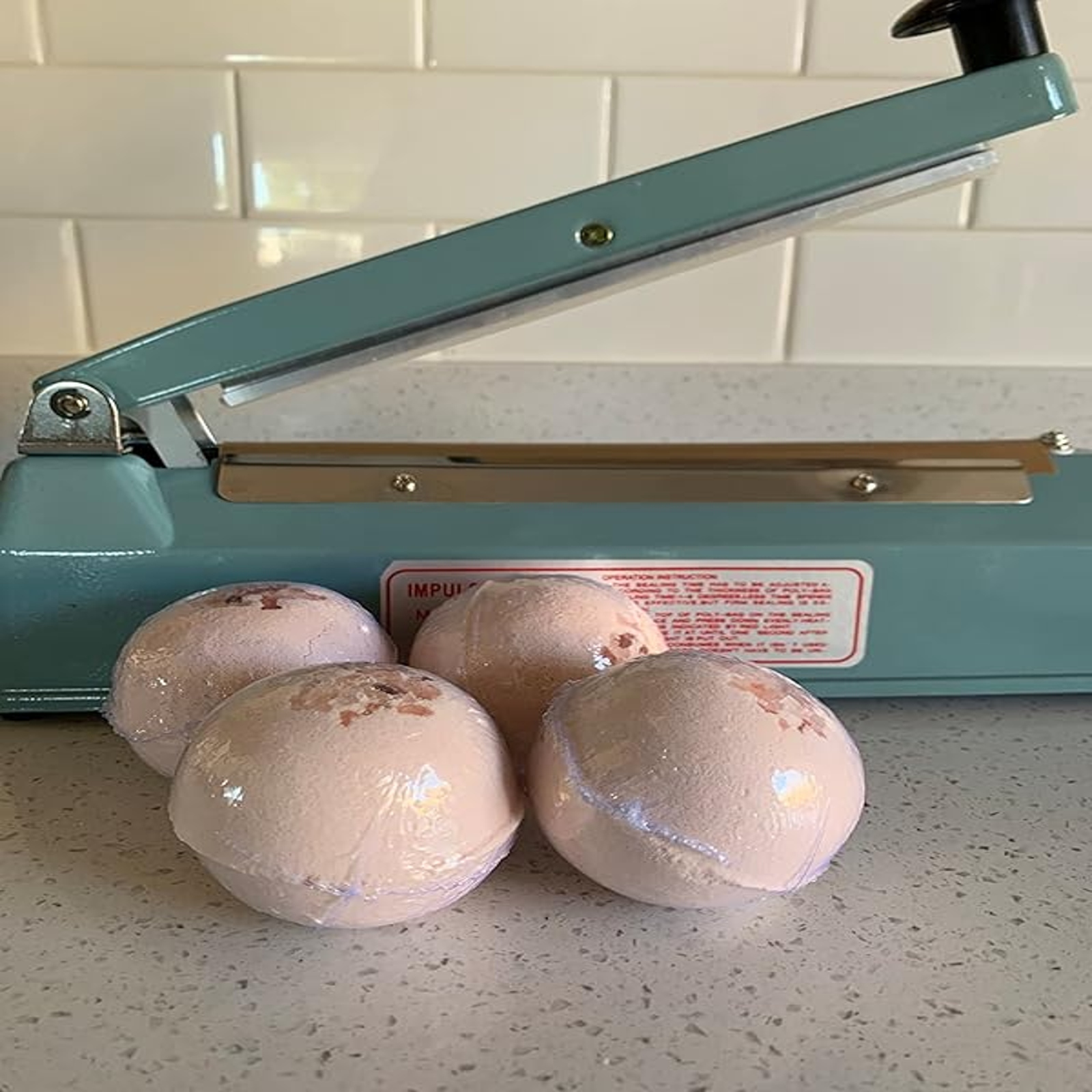} & 
\includegraphics[width=2.8cm, height=2.8cm]{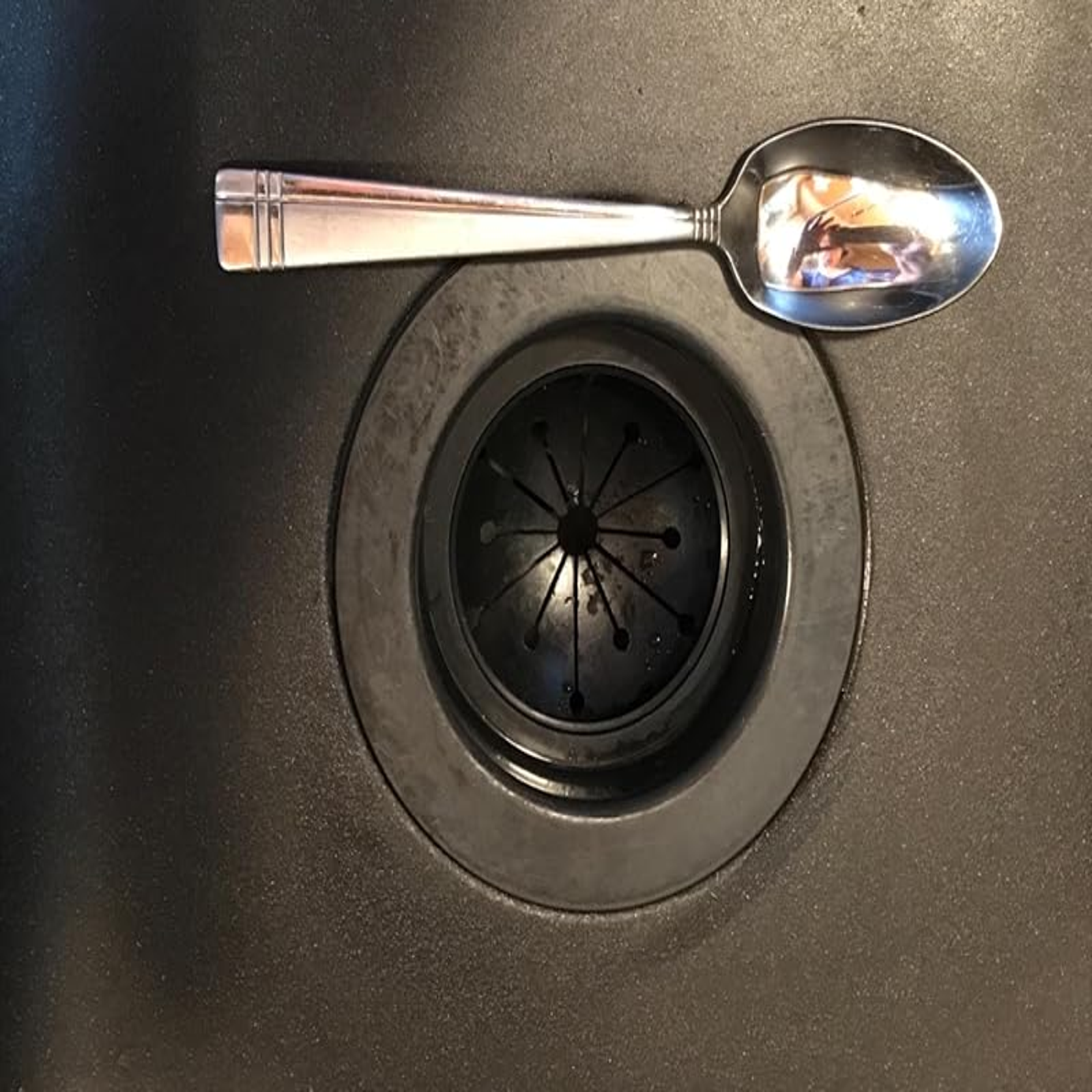} & 
\includegraphics[width=2.8cm, height=2.8cm]{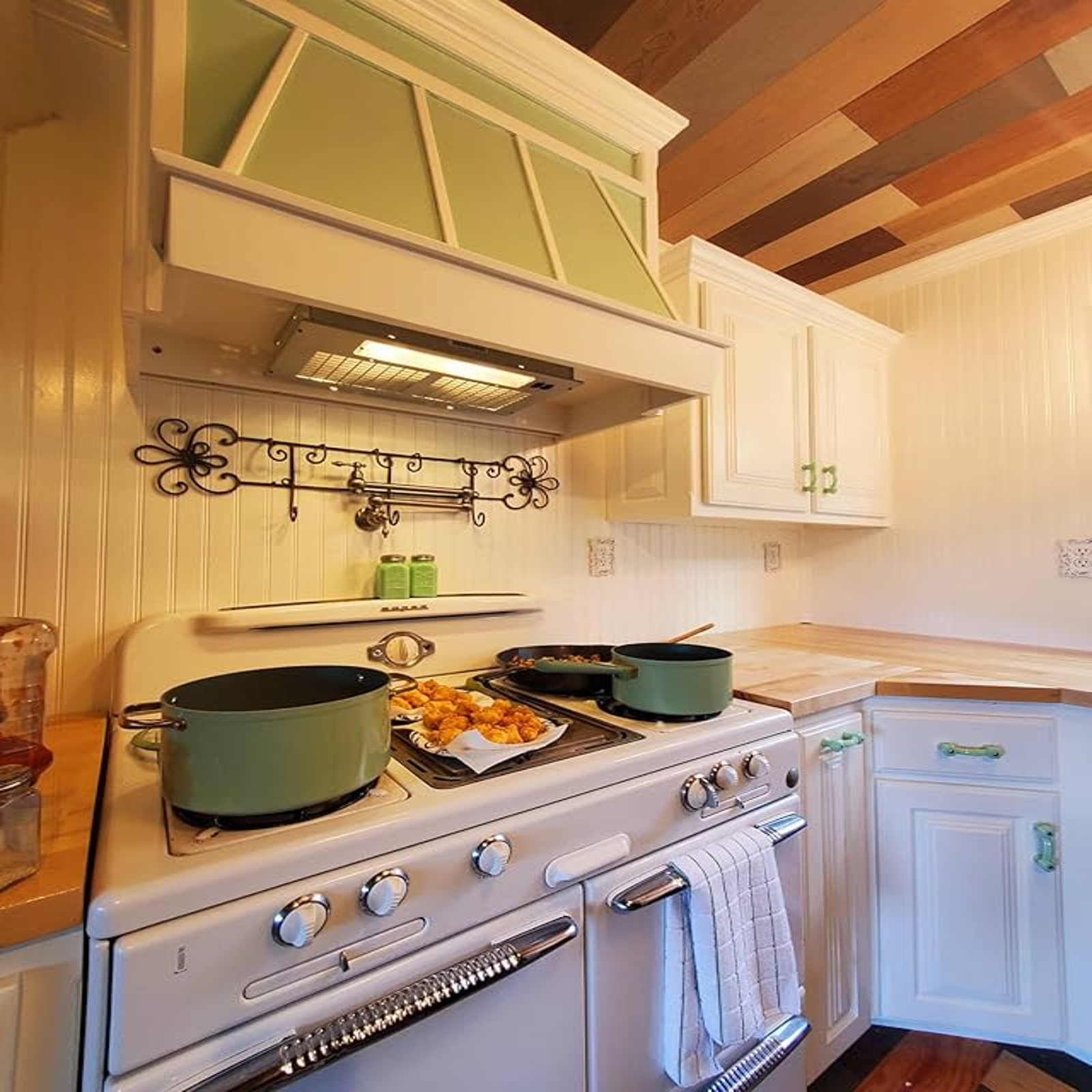} & 
\includegraphics[width=2.8cm, height=2.8cm]{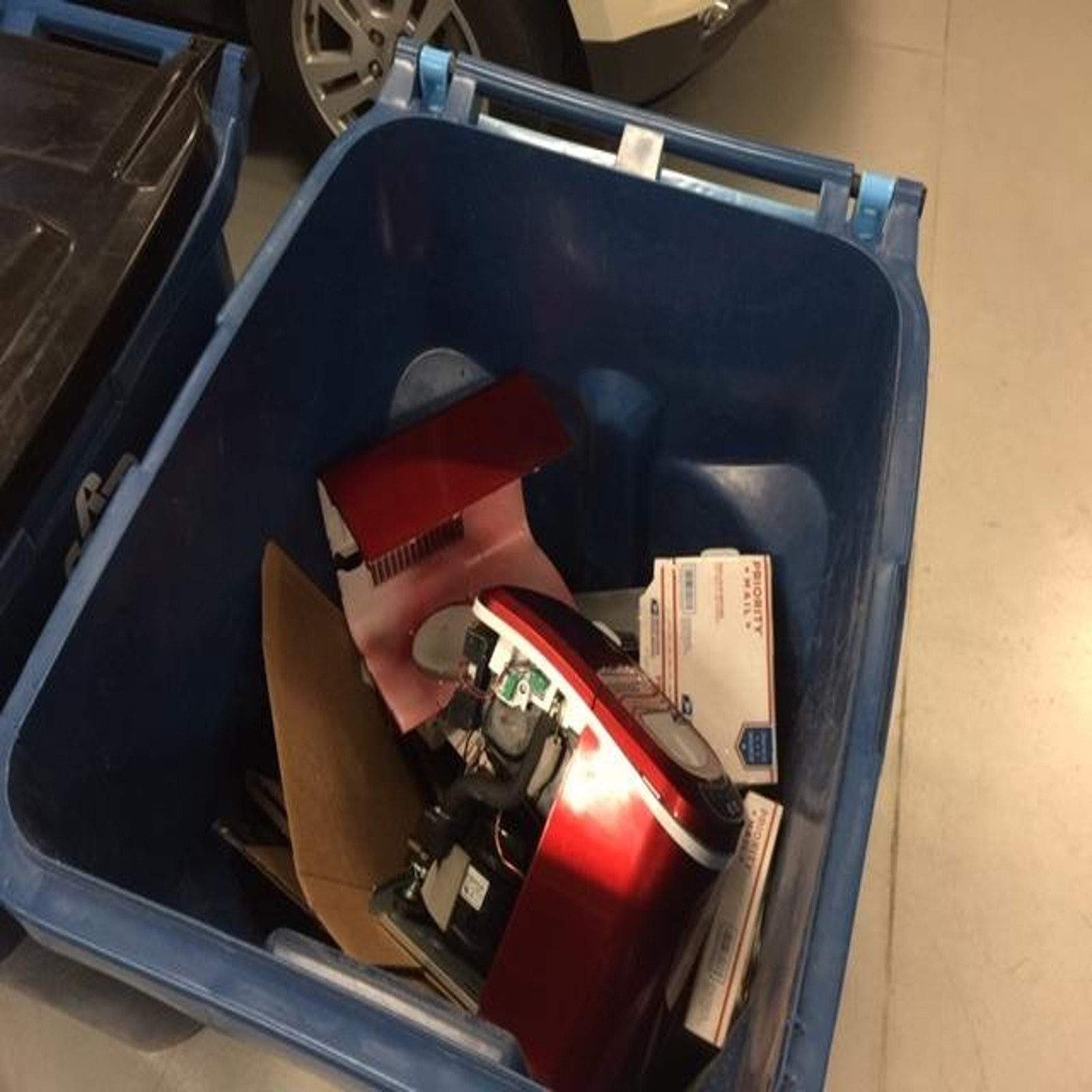} & 
\includegraphics[width=2.8cm, height=2.8cm]{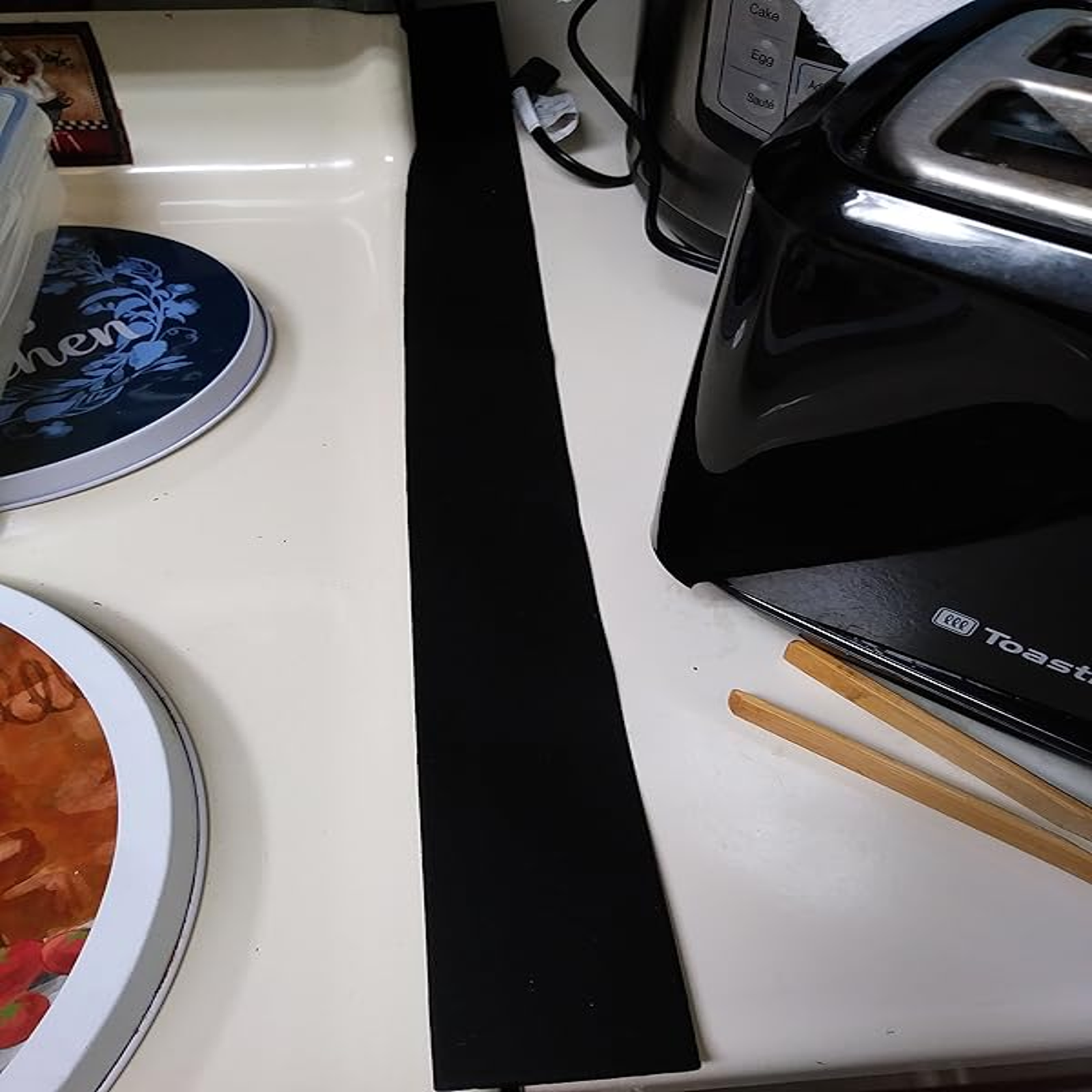}   \\ \hline

\multirow{2}{*}{\textbf{Actual Product}} & \multirow{2}{*}{Impulse Sealer} & \multirow{2}{*}{Garbage Disposal} & \multirow{2}{*}{Range Hood} & \multirow{2}{*}{Compact Ice Maker} & Silicone Stove Gap \\ 

& & & & & Cover \\ \hline

 & & & & & \\ [\dimexpr-\normalbaselineskip+0.5pt]
\textbf{Eigen CAM} & 
\includegraphics[width=2.8cm, height=2.8cm]{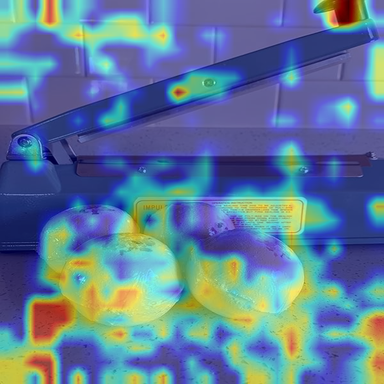} & 
\includegraphics[width=2.8cm, height=2.8cm]{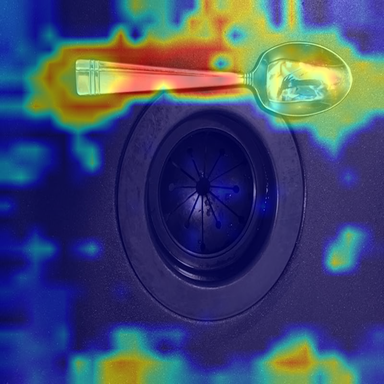} & 
\includegraphics[width=2.8cm, height=2.8cm]{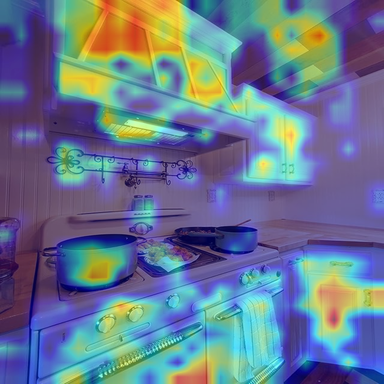} & 
\includegraphics[width=2.8cm, height=2.8cm]{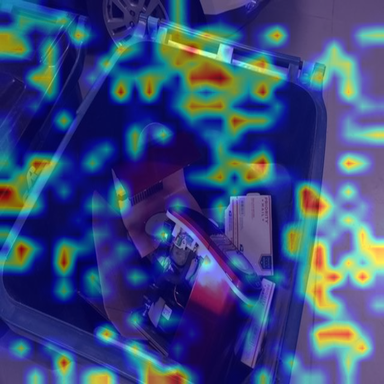} & 
\includegraphics[width=2.8cm, height=2.8cm]{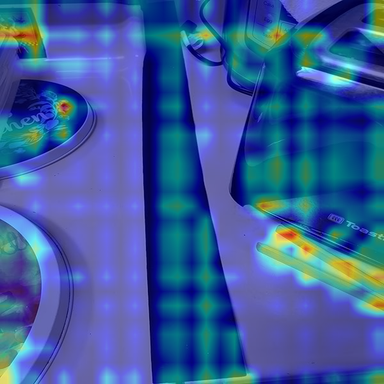}  \\ \hline

\multirow{2}{*}{\textbf{Predicted Product}} & \multirow{2}{*}{Bath bombs} & \multirow{2}{*}{Spoon} & \multirow{2}{*}{Oven} & \multirow{2}{*}{Shaver} & \multirow{2}{*}{Toaster} \\
& & & & & \\ \hline

\end{tabular}
\end{adjustbox}
\caption{\small Qualitative analysis of product identification.}
\label{fig:eigen_cam_comparison}
\end{figure}

Table \ref{tab:rating_metrics} presents performance of various CNN-based models and VLMs for the rating prediction task. LLaVa outperformed all in overall accuracy, achieving 32.90\%. It highlights the extreme inherent difficulty of extracting nuanced, subjective ratings purely from review images. Furthermore, the granular rating-wise breakdown reveals a pronounced bimodal bias across many models, which excel at predicting extreme ratings (1 or 5) but fail almost entirely to accurately classify moderate ratings (2, 3, and 4). Crucially, while advanced generalized VLMs like Qwen excelled in the prior product identification, they exhibit severe performance degradation here. This shows architectural divergence, where specialized or traditional models outperform generalized models on subjective tasks. This motivates modeling rating prediction as a dedicated task within the framework.
\noindent
\textbf{Qualitative Analysis:} 
Fig. \ref{tab:qualitative_results1}:(a) presents a qualitative analysis of the reviews generated by the proposed framework. The results demonstrate the architecture's capability to accurately recognize the prominent product and extract its associated visual attributes, including appearance, packaging, usability, size, and perceived quality. Furthermore, the generated text strictly adheres to the internally predicted rating, ensuring the production of coherent, sentiment-aware reviews. Additionally, Fig. \ref{fig:eigen_cam_comparison}:(a) shows the framework's precision in localizing and identifying products within the source images. By applying EigenCAM projections, we establish clear visual grounding and interpretability, providing qualitative insight into the spatial regions associated with product identification.


Fig. \ref{fig:eigen_cam_comparison}:(b) illustrates cases where the product identification agent incorrectly recognises the input product. These misidentifications propagate to the downstream stages, leading to inaccurate rating prediction and irrelevant review generation, as further shown in Fig.s \ref{tab:qualitative_results1}:(b)-(c). This highlights the error-propagation behaviour of the sequential multi-agent pipeline, where mistakes in early product grounding can negatively affect both sentiment estimation and final review generation.


\section{Conclusion}
\label{5sec:discussion_conclusion}
This work presents a multi-agent vision-language framework for generating product-specific and sentiment-aware reviews from e-commerce product images. By decomposing the task into product identification, rating prediction, visual evidence generation, and final review synthesis, the proposed architecture improves interpretability and provides better control over the review-generation process. The results show that the finetuned multi-agent pipeline produces coherent and visually grounded reviews, while agent-level evaluations confirm the importance of selecting suitable models for each stage.
However, the framework has some limitations. Since the system relies only on product images during inference, it cannot fully capture subjective user experiences such as durability, delivery issues, long-term usage, or functional defects. In addition, errors in product identification can propagate to rating prediction and final review generation, leading to irrelevant or mismatched reviews. Rating prediction also remains challenging because visual appearance alone may not reliably indicate user satisfaction.
Overall, the proposed approach demonstrates the potential of multi-agent vision-language systems for AI-assisted e-commerce review generation. Future work will focus on reducing error propagation, improving rating prediction, incorporating uncertainty-aware agent decisions, and extending the system with human-in-the-loop review refinement to ensure more reliable and trustworthy generated feedback.

\bibliographystyle{splncs04}
\bibliography{ref}





\end{document}